\documentclass[10pt,journal,compsoc]{IEEEtran}
\ifCLASSOPTIONcompsoc
  \usepackage[nocompress]{cite}
\else
  \usepackage{cite}
\fi
\ifCLASSINFOpdf
   \usepackage[pdftex]{graphicx}
   \usepackage{xcolor,tikz, subcaption, color}
\else
\fi
\usepackage{amsmath, amssymb, amsfonts, bm}

\usepackage[ruled, vlined, commentsnumbered, linesnumbered]{algorithm2e}
\usepackage{algorithmic}

\usepackage{color}
\usepackage{xcolor}

\usepackage{booktabs,tabularx}
\usepackage[symbol]{footmisc}

\usepackage{array, multirow}

\usepackage{url, multirow, hyperref}
\usepackage{kotex}

\newcommand{\etal}{{\em et al.}}       % et al.
\newcommand{\eg}{{\em e.g.}}           % e.g.
\newcommand{\ie}{{\em i.e.}}           % i.e.
\newcommand{\etc}{{\em etc}}         % etc.

\newcommand{\Exp}{\operatorname{Exp}}
\newcommand{\Log}{\operatorname{Log}}

\begin{document}
\title{Deep Efficient Continuous Manifold Learning \\for Time Series Modeling}

\author{Seungwoo Jeong,
        Wonjun Ko,
        Ahmad Wisnu Mulyadi,
        and~Heung-Il~Suk,~\IEEEmembership{Senior Member,~IEEE}% <-this % stops a space
        
\IEEEcompsocitemizethanks{
\IEEEcompsocthanksitem S. Jeong is with the Department of Artificial Intelligence, Korea University, Seoul 02841, Republic of Korea (e-mail: sw\_jeong@korea.ac.kr).
\IEEEcompsocthanksitem W. Ko and A. W. Mulyadi are with the Department of Brain and Cognitive Engineering, Korea University, Seoul 02841, Republic of Korea (e-mail: wjko@korea.ac.kr and wisnumulyadi@korea.ac.kr).
\IEEEcompsocthanksitem H.-I. Suk is with the Department of Artificial Intelligence and the Department of Brain and Cognitive Engineering, Korea University, Seoul 02841, Republic of Korea and the corresponding author (e-mail: hisuk@korea.ac.kr).
}% <-this % stops an unwanted space
% \thanks{Manuscript received April 19, 2005; revised August 26, 2015.}
}


\markboth{IEEE TRANSACTIONS ON PATTERN ANALYSIS AND MACHINE INTELLIGENCE}%
{Jeong \MakeLowercase{\textit{et al.}}: Efficient Continuous Manifold Learning for Time Series Modeling}


\IEEEtitleabstractindextext{%
\begin{abstract}
Modeling non-Euclidean data is drawing extensive attention along with the unprecedented successes of deep neural networks in diverse fields. Particularly, a symmetric positive definite matrix is being actively studied in computer vision, signal processing, and medical image analysis, due to its ability to learn beneficial statistical representations. However, owing to its rigid constraints, it remains challenging to optimization problems and inefficient computational costs, especially, when incorporating it with a deep learning framework. In this paper, we propose a framework to exploit a diffeomorphism mapping between Riemannian manifolds and a Cholesky space, by which it becomes feasible not only to efficiently solve optimization problems but also to greatly reduce computation costs. Further, for dynamic modeling of time-series data, we devise a continuous manifold learning method by systematically integrating a manifold ordinary differential equation and a gated recurrent neural network. It is worth noting that due to the nice parameterization of matrices in a Cholesky space, training our proposed network equipped with Riemannian geometric metrics is straightforward. We demonstrate through experiments over regular and irregular time-series datasets that our proposed model can be efficiently and reliably trained and outperforms existing manifold methods and state-of-the-art methods in various time-series tasks.
\end{abstract}

}

\maketitle

\IEEEdisplaynontitleabstractindextext

\IEEEpeerreviewmaketitle

\IEEEraisesectionheading{\section{Introduction}\label{sec:introduction}}
\IEEEPARstart{T}{ime} series data is a set of sequential observation values over time that is ubiquitous in the real-world such as video, natural language, audio, bio-signals, \etc. For example, in computer vision, a video is a set of consecutive images that await to be explored in diverse tasks such as action recognition \cite{zhen2019dilated}, object tracking \cite{zhang2021fairmot}, \etc. In natural language processing, a sequence of words is used for tasks of document classification \cite{devlin2018bert}, sentiment analysis \cite{devlin2018bert}, and language translation \cite{vaswani2017attention}. Furthermore, electroencephalogram (EEG) and electrocardiogram (ECG) data are bio-signals, frequently used for diagnostic-related tasks, such as sleep staging \cite{phan2021xsleepnet} and seizure detection \cite{ko2021multi}, respectively. Recently, with the development of devices, it has become possible to densely record more complex observations, and accordingly, more accurate and effective modeling of time-series data is essential.

\begin{figure}
    \centering
    \includegraphics[width=1\linewidth]{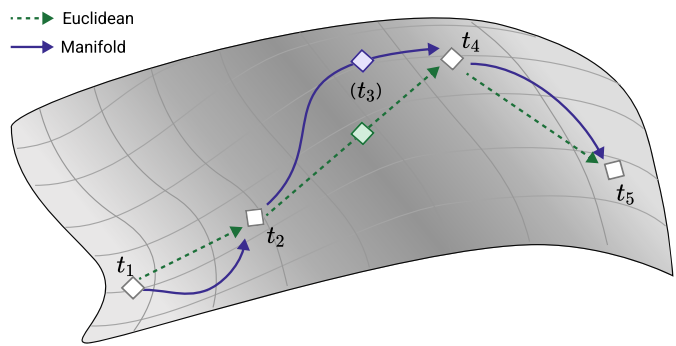}
    \caption{Depicted of difference between two sequential points in manifold space and Euclidean space. In particular, modeling and handling for the missing value $t_3$ require a technique that considers the characteristics of the manifold space.}
    \label{fig:fig_latentspace}
\end{figure}

\begin{table*}[th]
    \centering
    \caption{Summary of related works categorized by space, continuous, approach.}
    \label{table1}
    \begin{tabular}{cccccc}
        \toprule
        Methods & Space & Continuous & Irregularity & Description \\  % Approach
        \midrule
        GRU-D~\cite{che2018recurrent} & & & $\checkmark$ & GRU-based missing pattern representation learning \\ % RNN
        WEASEL+MUSE~\cite{schafer2017multivariate} & \multicolumn{1}{c}{\multirow{7}{*}{Euclidean}} & & & Similarity measure with bag-of-pattern \\ %Distance
        TapNet~\cite{zhang2020tapnet} & & & & Distance-based low-dimensional feature representation learning \\ %CNN + RNN
        ShapeNet~\cite{li2021shapenet} & & & & Shapelet-neural network \\ % CNN
        TS2Vec \cite{yue2022ts2vec} & & & & Hierarchical contrastive learning for a universal time-series representation \\ 
        ODE-RNN~\cite{rubanova2019latent} & & $\checkmark$ & $\checkmark$ & Learning hidden dynamics by neural ODE \\ %ODE
        GRU-ODE~\cite{de2019gru, jordan2021gated} & & $\checkmark$ & $\checkmark$ & Neural ODE analogous to GRU \\ % RNN
        ODE2VAE~\cite{yildiz2019ode2vae} & & $\checkmark$ & $\checkmark$ & Second order ODE for high-dimensional continuous data \\ %ODE
        Neural CDE~\cite{kidger2020neural} & & $\checkmark$ & $\checkmark$ & Neural controlled differential equation model for irregularly-sampled data \\ % CDE
        \midrule
        SPDSRU~\cite{chakraborty2018statistical} & \multicolumn{1}{c}{\multirow{3}{*}{Manifold}}  & & &  Arithmetic and geometric operations for SPD hard constraints \\ % RNN
        ManifoldDCNN~\cite{zhen2019dilated} & & & & Dilated CNN network for manifold-valued sequential data \\ % CNN
        MCNF \cite{lou2020neural, falorsi2020neural} & & $\checkmark$ & & Neural manifold ODE for continuous normalizing flows \\ % CNN
        \midrule
        \multirow{2}{*}{Ours} & \multirow{2}{*}{Manifold} & \multirow{2}{*}{$\checkmark$} & \multirow{2}{*}{$\checkmark$}  & Systematic integration of GRU and ODE \\
        & & & & with diffeomorphism mappings from SPD space to Cholesky space \\ %ODE 
        \bottomrule
    \end{tabular}
\end{table*} 

While there has been a number of literature on algorithms and pattern analysis methods, deep learning has recently been most exhaustively employed as the de facto standard for time-series modeling. These approaches use convolutional neural networks (CNNs) or recurrent neural networks (RNNs) to capture spatial or temporal features of time-series data \cite{yang2017tensor, karim2019multivariate, franceschi2019unsupervised, zhang2020tapnet, li2021shapenet}. 
Recently, Transformer-based methods \cite{zhou2021informer, zerveas2021transformer} have been used to handle the lack of long-range dependency, which is one of the problems of RNN.
However, these neural network-based methods suffer from (i) an inadequacy in handling non-Euclidean characteristics in data \eg, processing data in manifold space and (ii) insufficient data for processing complex and high-dimensional data.  
In this context, utilizing manifold space for modeling in deep learning for time series data offers several notable advantages compared to traditional Euclidean approaches. By effectively capturing the intrinsic non-linearity inherent in time series data and employing a mapping procedure onto lower-dimensional manifold, deep learning models become capable of adeptly capturing intricate temporal patterns and dependencies that would otherwise go unnoticed in Euclidean representations \cite{li2023combination, ding2023black, chen2017deep}. Consequently, this facilitates the attainment of more precise and expressive data representations. Furthermore, the adoption of manifold space learning engenders heightened robustness of models to the influences of noise and outliers \cite{ding2023black}, an augmented capacity for generalization through an emphasis on crucial variations \cite{chen2017deep}, hence providing a more condensed and abstract perspective of temporal patterns. These advantages collectively contribute to advancing more dependable, efficient, and interpretable models for time series analysis.

Simultaneously, researchers have actively explored manifold space as a computational method for efficiently handling the geometric features of data.
Such approaches are often used for the analysis of various representations, including but not limited to shapes \cite{chang2015shapenet}, graphs and trees \cite{scarselli2008graph, kipf2016semi}, symmetric positive definite (SPD) matrices \cite{moakher2005differential, jayasumana2013kernel}, and rotation matrices \cite{kendall2017geometric}.
Of these data types, the SPD matrix has received considerable attention in a variety of application areas due to its ability to learn suitable statistical representations \cite{huang2017riemannian, suh2021riemannian, chakraborty2020manifoldnet}. In computer vision, SPD matrices have achieved great success as representations for classification tasks such as skeletal-based human behavioral recognition \cite{nguyen2019neural} and emotion recognition \cite{huang2017riemannian}. In signal processing, it has also been used as an expression method for multivariate time-series (MTS) data \cite{suh2021riemannian}. In neuroimage data analysis, a diffusion tensor image, which indicates structural connections over brain regions, has been treated as manifold-valued data \cite{chakraborty2020manifoldnet}.
Nevertheless, the direct use of SPD matrices remains a challenge because of its rigid constraints. SPD matrices in the Euclidean metric cause unfavorable results, such as the swelling effect \cite{chefd2004regularizing} and a finite distance of symmetric matrices with non-positive eigenvalues \cite{arsigny2007geometric,lin2019riemannian}. Thus, it is unreasonable for conventional deep-learning methods to model SPD matrices by ignoring their geometric characteristics and computational constraints.
% and for training deep learning models, it is still a challenging problem that the hard constraint of the SPD matrix. 
Additionally, backpropagation in the optimization process makes it difficult for operand components within deep neural networks to maintain positive definiteness. To address these problems, several methods have been proposed to train the model while maintaining the structure of the SPD matrix. For instance, SPDNet \cite{huang2017riemannian} handles this issue through an orthogonal weight matrix on the compact Stiefel manifolds. In \cite{gao2020learning}, the optimizer structure was designed through a meta-learning-based method. Furthermore, existing deep learning methodologies, \eg, CNN-based models \cite{zhen2019dilated, chakraborty2020manifoldnet} or RNN-based models \cite{chakraborty2018statistical}, were proposed to deal with the manifold-valued sequential data, suggesting a recursive method for the weighted Fr\'echet mean to avoid the problems of high computational cost and instability in optimization.
However, numerical errors and excessive costs remain challenges in the computation process.

Generally, modeling time-series data is based on the premise of discrete observations and emission intervals. However, data in the real world are mainly irregularly-sampled data possibly due to unexpected errors in recording or device. 
Fig. \ref{fig:fig_latentspace} presents an example of time-series data with one missing value at a time point $t_{3}$ on a discrete-time interval. When modeling and handling the missing value using an imputation technique, \eg, bi-linear interpolation, in a Euclidean metric (green-colored arrows), it causes a large error compared to the one estimated in a manifold metric. Even when given two data points with the same interval, its latent representations in the manifold take the form of a curve (rather than a linear form) due to the geometric characteristics. Notably, it is beneficial to model in a continuous-time space so that the geometric features inherent in the data are better represented, thus improving target downstream tasks. Therefore, it is of paramount importance to develop a continuous manifold learning method for time-series modeling.

A recent provocative approach to deal with irregularly sampled data is a continuous-time model through ordinary differential equations (ODE) with a time variable \cite{chen2018neural}. Neural ODE \cite{chen2018neural} and its variants \cite{rubanova2019latent, de2019gru} have demonstrated their capability in signal representation by capturing dynamics over the internal hidden states.
Subsequent studies extend the neural ODE to the manifold space \cite{lou2020neural, falorsi2020neural}. For example, Manifold ODE \cite{lou2020neural} has been proposed to learn the dynamics of the states in the Riemannian manifold, aiming to estimate complicated distributions in the framework of normalizing flow \cite{rezende2015variational}. Although various neural ODEs for manifold space have been proposed, their application for time-series data is still limited and, even if applicable, computationally expensive and inefficient.

In this work, we propose a novel method of continuous manifold learning on sequential manifold-valued data for time-series modeling. Based on rigorous mathematical justifications, we propose to exploit a diffeomorphism mapping between Riemannian manifolds and a Cholesky space \cite{lin2019riemannian}, by which it becomes feasible to efficiently solve optimization problems and to greatly reduce computational costs. We devise an algorithm to solve an optimization problem under the SPD constraint and achieve an effective computation cost by re-defining operations within a deep neural network framework. Our proposed method represents manifold-valued sequential data via a practical computation based on well-defined mathematical formulations. Particularly, we introduce a recurrent neural model on the Cholesky space, making it possible to greatly enhance computational efficiency and train network parameters with conventional backpropagation. Moreover, the proposed method successfully learns real-world continuous data modeling for irregularity or sparsity and adjusts the discrepancy in geodesics with the use of a manifold ODE for dynamics representation utilizing an RNN-based network. To prove the validity and robustness of our proposed method, we conducted exhaustive experiments and achieved promising performance over various public datasets.

In summary, there are three major contributions of this work:
\begin{itemize}
    \item To handle the hard constraints of an SPD matrix in time-series modeling, we devise a novel computational mechanism that exploits a diffeomorphism mapping from SPD space to Cholesky space and applies it to a deep recurrent network. Our proposed method greatly reduces computational costs and numerical errors, thus enabling robust and stable learning.
    % To handle the hard constraints of a SPD matrix {\HI in learning}, we devise a novel recurrent model that exploits a diffeomorphism mapping from SPD space to Cholesky space, by which we can greatly reduce computational costs as well as numerical errors, thus enabling stable training and robust representations. % By doing so, it prevents numerical errors, reduces computational costs, and enables stable training.
    \item We also develop a manifold ODE representing temporal dynamics inherent in time-series data, allowing it to learn trajectory with geometric characteristics of sequential manifold-valued data.
    \item We exhaustively demonstrate the effectiveness of the proposed method through experiments over various public datasets, achieving state-of-the-art performance.
\end{itemize}
We organized the rest of this work as follows. Section \ref{sec:related work} briefly reviews related work. In Section \ref{sec:preliminaries}, we describe essential preliminary concepts to pave the way for understanding the proposed method. Then, Section \ref{sec:Method} introduces the proposed method for continuous manifold learning with the diffeomorphism estimation. Section \ref{sec:experiments} describes experimental settings in detail for reproducibility and results are compared with existing methods. Finally, we conclude the study and recommend future research directions in Section \ref{sec:conclusion}.

\begin{table*}[th]
\centering
\caption{Summary of operations in the Riemannian space and the Cholesky space. The notations used in the operations are as follows; $d$: the dimension of a matrix, $P$: a point in Riemannian space, $Q$: a point in tangent space of Riemannian space, $X, Y, L$: points in Cholesky space, $K$: a point in tangent space of Cholesky space. Note that for the distance in Riemannian space, we denote the affine-invariant Riemannian metric.}
\label{table2}
\begin{tabular}{ccc}
\toprule
\multirow{2}{*}{Manifold} & Riemannian space & Cholesky space \\
 &  $\mathcal{S}^+_d = \{ P \in \mathbb{R}^{d\times d}: P = P^\top,\mathbf{x}^\top P \mathbf{x} > 0, \forall \mathbf{x} \in \mathbb{R}^{d}\setminus\{\mathbf{0}_d\}\}$ & $\mathcal{L}_+ = \{X \in \mathbb{R}^{d\times d}: \mathcal{D}(X)>0, \{X_{ij}\}_{i>j} = 0\}$ \\ \hline
Metric & $\mathcal{T}_X\mathcal{M} \times \mathcal{T}_X\mathcal{M}$  & $\sum_{i>j}X_{ij}Y_{ij} + \sum^m_{j=1}X_{jj}Y_{jj}L^{-2}_{jj}$ \\ \hline
Distance & $\frac{1}{2}||\log(P_1^{-\frac{1}{2}}P_2P_1^{-\frac{1}{2}})||_F$ & $ \{||\lfloor L \rfloor - \lfloor K\rfloor||^2_F + ||\log\mathcal{D}(L) - \log\mathcal{D}(K)||^2_F\}^{\frac{1}{2}}$ \\ \hline
Exponential map & $P^{\frac{1}{2}}\exp(P^{\frac{1}{2}}QP^{-\frac{1}{2}})P^{\frac{1}{2}}$ & $\lfloor X\rfloor + \lfloor K\rfloor + \mathcal{D}(X)\exp\{\mathcal{D}(K)\mathcal{D}(X)^{-1}\}$ \\ \hline
Logarithmic map & $Q^{\frac{1}{2}}\log(Q^{\frac{1}{2}}PQ^{-\frac{1}{2}})Q^{\frac{1}{2}}$ & $\lfloor X\rfloor - \lfloor K\rfloor + \mathcal{D}(K)\log\{\mathcal{D}(K)^{-1}\mathcal{D}(X)\}$ \\ \bottomrule
\end{tabular}
\end{table*}

\section{Related Work}\label{sec:related work}
In this section, we introduce existing research on discrete and continuous time-series modeling.  TABLE \ref{table1} compares recent methods with our proposed in terms of computing space, continuous, and time regularity.
\subsection{Discrete Time-Series Modeling}
Most general models employed as baselines in time-series data are distance function-based methods \cite{bagnall2018uea}.
Traditionally, nearest neighbor (NN) classifiers with distance functions \cite{lines2015time} have been widely used, and the method used with Dynamic Time Warping (DTW) in particular has shown good performance \cite{ruiz2021great}. In WEASEL-MUSE \cite{schafer2017multivariate}, a bag of Symbolic Fourier Approximation (SFA) is used for MTS classification. However, these classical models have limited scope and may face challenges when dealing with complex patterns, as well as limitations in terms of computational cost, sensitivity to outliers, and potential data loss.

Recently, deep learning-based methods have shown promising performance. In MLSTM-FCN \cite{karim2019multivariate}, a model combining long short-term memory (LSTM) and CNN is proposed for MTS classification tasks. Specifically, performance is improved by adding squeeze-and-excitation blocks \cite{hu2018squeeze}.
In \cite{franceschi2019unsupervised}, an unsupervised method is proposed to solve the variable lengths and sparse labeling problems of time-series data. The authors process the variable length through a dilated convolutional \cite{oord2016wavenet} encoder and build a triplet loss \cite{schroff2015facenet} with negative samples. Meanwhile, for high-dimensional multivariate data and limited training label problems, TapNet \cite{zhang2020tapnet} proposes an attentional prototype network by learning low-dimensional features through random group permutation.
In ShapeNet \cite{li2021shapenet}, the author proposes an MDC-CNN with a triplet loss that trains time-series subsequences of various lengths in a unified space. Meanwhile, transformer-based methods \cite{zhou2021informer, zerveas2021transformer} have been mainly used for forecasting time-series data, especially due to their ability to model long-range dependencies.
Representation learning for time series data has garnered significant attention in recent years. Of various techniques, contrastive learning-based methods have emerged as prominent approaches, offering universal representation learning for various time series data tasks \cite{yue2022ts2vec, zerveas2021transformer}. Zerveas \etal~\cite{zerveas2021transformer} propose an unsupervised representation learning method based on transformers, evaluating its effectiveness in multivariate time series regression and classification. In \cite{yue2022ts2vec}, TS2Vec is introduced as a technique for learning universal representations at different semantic levels.
Although these models use deep learning-based methods to learn spatio-spectral-temporal features from time-series data, they are limited in the assumption that the data have constant intervals and thus are most suitable for regular time-series data. In contrast, our proposed method is a neural ODE-based method that enables continuous time-series modeling.
% {\sw However, these methods have limitations in handling sparse or irregularly-sampled data. On the other hand, our proposed method is a neural ODE-based method that enables continuous time-series modeling.}

\subsection{Continuous Modeling} %Time-Series Modeling
Meanwhile, researchers have focused on modeling dynamic patterns at continuous time points.
Foremost, \cite{chen2018neural} proposes the idea of parameterizing derivatives of hidden states for continuous modeling.
Inspired by neural ODEs, \cite{rubanova2019latent} propose latent ODE and ODE-RNN to deal with irregularly-sampled data.
In \cite{de2019gru}, GRU-ODE-Bayes, combining GRU-ODE and GRU-Bayes, is proposed to address sporadic observations that occur in multidimensional time-series data.
Theoretically similar to \cite{de2019gru}, in Neural CDE \cite{kidger2020neural}, an extended model of Neural ODE using a controlled differential equation is proposed to address the limitations of an ordinary differential equation.
Meanwhile, in ODE2VAE, by decomposing the latent space into position and velocity components, time-sequential modeling for sequential data is proposed \cite{yildiz2019ode2vae}.
In \cite{chen2021continuous}, for the relevance or importance of individual samples, a continuous-time attention method is proposed by applying the attention mechanism to the neural ODE.

% Recent studies on the use of neural ODEs have begun to model on structured \cite{xhonneux2020continuous} or manifold-valued data \cite{lou2020neural}. In order to characterize the continuous node representations, \cite{xhonneux2020continuous} propose a graph-based model. There are also recent studies on normalizing flow using neural ODEs in manifold space \cite{lou2020neural, falorsi2020neural}. Despite the high-dimensional characteristics of recent time-series data, the aforementioned continuous models do not consider time-series modeling in the manifold space. In contrast, our model uses neural ODEs by focusing on continuous time-series modeling for the trajectories of hidden states with structural features in manifold space.
Recent studies on the use of neural ODEs have begun to model structured \cite{xhonneux2020continuous} or manifold-valued data \cite{lou2020neural}. To capture the essence of continuous node representations, \cite{xhonneux2020continuous} introduces a graph-based model. Within the research community, neural ODEs have been expanded from their original domain of Euclidean space to encompass a manifold space, particularly in normalizing flow tasks \cite{lou2020neural, falorsi2020neural}. Even with the potential demonstrated by these advancements, a substantial gap exists in research regarding applying neural ODEs to high-dimensional time-series data. In contrast, our model takes a different approach by employing neural ODEs to specifically address continuous time-series modeling, focusing on the trajectories of hidden states within a manifold space while incorporating structural features.

\subsection{Manifold Learning}
In order to better represent the geometric characteristics inherent in time-series data, recent studies focus on modeling with manifold-valued data, especially SPD matrices, in deep learning frameworks.
In \cite{huang2017riemannian}, the author proposes an SPDNet that consists of fully connected convolution-like layers, rectified linear units-like layers, and eigenvalue logarithm layers to train the deep model while maintaining the data structure on the SPD matrix. Subsequently, \cite{brooks2019riemannian} redefines the batch normalization on SPDNet using Riemannian operations to improve the classification performance and robustness of the model.
In \cite{suh2021riemannian}, the author proposes a metric learning-based SPD model that divides the classification problem into sub-problems and builds a single model to solve each sub-problem. Succinctly, SPDNet models are presented based on mapping with sub-manifold, but the repeated eigenvalue decomposition in the process inevitably generates tremendous computation costs.

Meanwhile, \cite{chakraborty2018statistical} proposes a deep model based on statistical machine learning when data are ordered, longitudinal, or temporal in a Riemannian manifold. Additionally, to resolve the slow speed issue of RNN-based models, the dilated and causal CNNs were defined in the manifold space \cite{chakraborty2020manifoldnet, zhen2019dilated}. In \cite{zhen2021flow}, a flow-based generative model for manifold-valued data is introduced through three types of invertible layers in a generative regime.
Training a deep model while maintaining the SPD constraint is challenging due to huge computational costs and unstable training.

Our approach differs from these previous SPD manifold methods, which directly process the SPD matrix. We accomplished this through mapping into the Cholesky space, which is easy to compute. This way, our proposed method relaxes the constraints enough to allow the optimization to be processed with the training algorithms.

\section{Preliminaries}\label{sec:preliminaries}
In this section, we concisely provide useful definitions and operations for Riemannian geometry.
The notations and operations in Riemannian and Cholesky spaces are summarized in TABLE \ref{table2}.
% \footnote{See Supplementary A to compare basic properties of Riemannian and Cholesky manifolds.}. 
Note that our development is based on the diffeomorphism mapping of Cholesky decomposition, thus in Riemannian Cholesky space rather than in the Riemannian geometry of SPD matrices. %$n \times n$ 
In the following, $\lfloor \cdot \rfloor$ and $\mathcal{D}(\cdot)$ denote the strictly lower triangular and the diagonal part of a matrix, respectively.

%% \subsection{Riemannian Geometry}
%% A Riemannian manifold $(\mathcal{M}, g)$ is a real, smooth manifold $\mathcal{M}$ equipped with a positive-definite inner product $g_\mathbf{x}$ on the tangent space $\mathcal{T}_\mathbf{x}\mathcal{M}$ at each point $\mathbf{x} \in \mathcal{M}$. The $g_\mathbf{x}$ is called a Riemannian metric, which makes it possible to define several geometric notions on a Riemannian manifold, such as angle, length of a curve.

%% \paragraph{Riemannian Manifold and Metric on Cholesky Space \cite{lin2019riemannian}.}

\begin{figure*}
  \centering
    \includegraphics[width=.9 \linewidth]{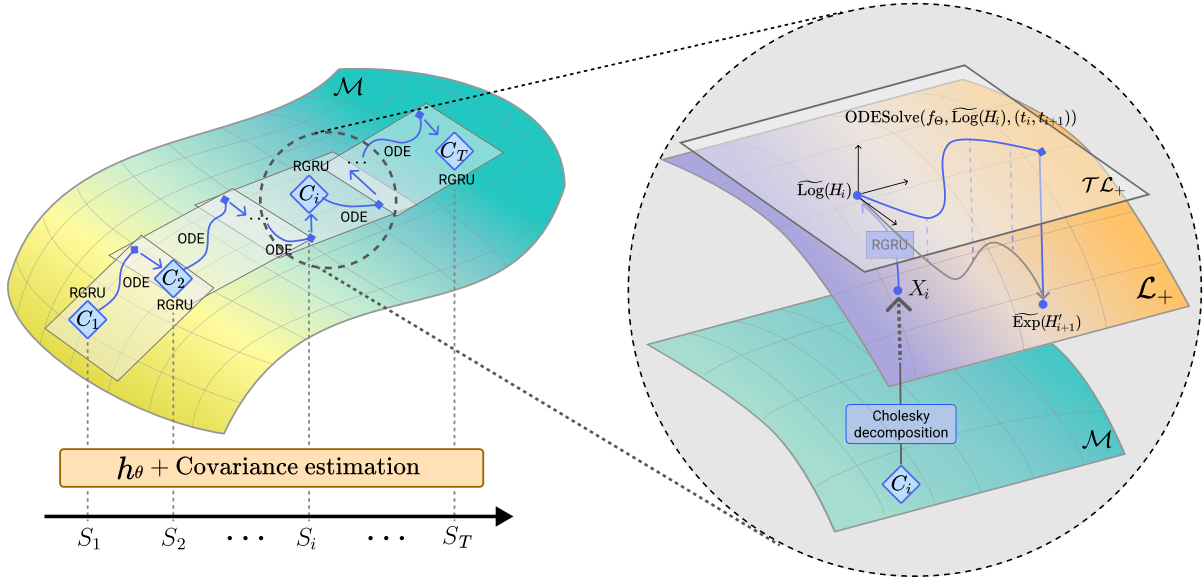}
    \caption{
    For the overall framework of the proposed method, first, input sequences $\{S_i\}_{1, \cdots, T}$ are fed into a feature extraction network $h_\theta$ for high-level representation. The manifold point $\{C_i\}_{1, \cdots, T}$ obtained by covariance estimation is then decomposed to $\{X_i\}_{1,\cdots, T}$ in the Cholesky space. Our proposed Riemannian manifold gated recurrent unit (RGRU) calculates Cholesky points, then the estimated points are mapped into the tangent space of the ordinary differential equation (ODE) solver ($\operatorname{ODESolve}$) through a logarithmic map $\widetilde{\operatorname{Log}}(\cdot)$. Afterward, the output of the ODE solver at the $i$-th timepoint $H'_i$ is projected to Cholesky space $\mathcal{L}_+$ by an exponential map $\widetilde{\operatorname{Exp}}(\cdot)$.}
    \label{fig_framework}
\end{figure*}

\subsection{Riemannian Manifold and Metric on Cholesky Space \cite{lin2019riemannian}.}
A Riemannian manifold $(\mathcal{M}, g)$ is a real, smooth manifold $\mathcal{M}$ equipped with a positive-definite inner product $g_X$ on the tangent space $\mathcal{T}_X\mathcal{M}$ at each point $X\in \mathcal{M}$. The $g_X$ is called a Riemannian metric, making it possible to define several geometric notions on a Riemannian manifold, such as the angle and curve length.

The space $\mathcal{S}^+_d$ of $d \times d$ SPD matrices is a smooth submanifold of the space $\mathcal{S}_d$ of symmetric matrices, whose tangent space at a given SPD matrix is identified with $\mathcal{S}_d$. 
Cholesky decomposition represents a matrix $S\in\mathcal{S}^{+}_{d}$ as a product of a lower triangular matrix $L$ and its transpose, \ie, $S=LL^{\top}$.
The smooth submanifold $\mathcal{L}_{+}$ of the space of lower triangular matrices $\mathcal{L}$ with diagonal elements all positive defines a Cholesky space.
%% Similarly, the space $\mathcal{L}_+$ called Cholesky space is a smooth submanifold of $\mathcal{L}$. 
Note that the tangent space of $\mathcal{L}_+$ at a given matrix $L \in \mathcal{L}_+$ is identified with the linear space $\mathcal{L}$.
Let the \textit{strict} lower triangular space $\lfloor \mathcal{L} \rfloor = \{\lfloor X \rfloor | X \in \mathcal{L}\}  \in \mathbb{R}^{d(d-1)/2}$ with the Frobenius inner product $\langle X, Y\rangle_F = \sum^d_{i,j=1}X_{ij}Y_{ij}$, $\forall$ $X, Y \in \lfloor\mathcal{L}\rfloor$, while denoting the diagonal part $\mathcal{D}(\mathcal{L}) = \{\mathcal{D}(X)|X \in \mathcal{L}\}$ with a different inner product $\langle \mathcal{D}(L)^{-1}\mathcal{D}(X),\mathcal{D}(L)^{-1}\mathcal{D}(Y)\rangle_F$.
Then, we define a metric $\Tilde{g}$ for the tangent space $\mathcal{T}_{L}\mathcal{L}_{+}$ as follows,
% \begin{align}
%     \begin{split}
%     \Tilde{g}_{\HI L}(X, Y) &= \langle X, Y\rangle_F + \langle \mathcal{D}(L)^{-1}\mathcal{D}(X),\mathcal{D}(L)^{-1}\mathcal{D}(Y)\rangle_F
%     \\ &=\sum_{i>j} X_{ij}Y_{ij} + \sum^d_{j=1}X_{jj}Y_{jj}L^{-2}_{jj}.
%     \end{split}
%     \label{eq1}
% \end{align}
\begin{align}
    \Tilde{g}_L(X, Y) &= \langle X, Y\rangle_F + \langle \mathcal{D}(L)^{-1}\mathcal{D}(X),\mathcal{D}(L)^{-1}\mathcal{D}(Y)\rangle_F \nonumber\\ 
    &=\sum^d_{i=1}\sum^{i}_{j=1} X_{ij}Y_{ij} + \sum^d_{j=1}X_{jj}Y_{jj}L^{-2}_{jj}.
    \label{eq1}
\end{align}
% Now, Cholesky space can be defined as $(\mathcal{L}_+, \Tilde{g})$.
% As a manifold map between smooth manifolds $\mathcal{L}_+$ and $\mathcal{S}^+_d$, the map $\mathfrak{L}$ is a diffeomorphism.
\vspace{5pt}\noindent\textbf{Proposition 1.} \cite{lin2019riemannian} \textit{Cholesky map $\mathfrak{L}$ is a diffeomorphism between smooth manifolds $\mathcal{L}_+$ and $\mathcal{S}^+_d$.}
\\

\noindent The full proof of Proposition 1 can be found in \cite{lin2019riemannian}.
Essentially, the Cholesky map easily represents deep learning processes in the SPD manifold space.

\subsection{Exponential and Logarithmic Maps}
For any $P \in \mathcal{S}^+_d$ and $Q \in \mathcal{S}_d$, the exponential map $\Exp_P:\mathcal{S}_d \rightarrow \mathcal{S}^+_d$ and the logarithmic map $\Log_Q:\mathcal{S}^+_d \rightarrow \mathcal{S}_d$ are defined as
\begin{align}
    \label{eq2}
    %\forall S \in \mathcal{S}_d, 
    \Exp_P(Q) &= P^{\frac{1}{2}}\exp(P^{\frac{1}{2}}QP^{-\frac{1}{2}})P^{\frac{1}{2}} \in S^+_d, \\ 
    \label{eq3}
    %\forall P \in \mathcal{S}^+_d, 
    \Log_Q(P) &= Q^{\frac{1}{2}}\log(Q^{\frac{1}{2}}PQ^{-\frac{1}{2}})Q^{\frac{1}{2}} \in S_d.
\end{align}
Generally, it is difficult to compute Riemannian exponential and logarithmic maps because they require evaluating a series of infinite number of terms \cite{arsigny2007geometric}. 
However, by mapping to Cholesky space, we obtain easy-to-compute expressions for Riemannian exponential and logarithmic maps. 
Let $X \in \mathcal{L}_+$ and $K \in \mathcal{L}$. Then, we define the exponential map $\widetilde\Exp$ and the logarithmic map $\widetilde\Log$ as in \cite{lin2019riemannian}
\begin{align}
    \label{eq4}
    \widetilde\Exp_X(K) &= \lfloor X\rfloor + \lfloor K\rfloor + \mathcal{D}(X)\exp\{\mathcal{D}(K)\mathcal{D}(X)^{-1}\}, \\
    \label{eq5}
    \widetilde\Log_K(X) &= \lfloor X\rfloor - \lfloor K\rfloor + \mathcal{D}(K)\log\{\mathcal{D}(K)^{-1}\mathcal{D}(X)\}.
\end{align}
For the convenience of expression, we omit the subscript (\ie, $\widetilde\Exp_X = \widetilde\Exp$)

\subsection{Fr\'echet Mean}
The Fr\'echet mean \cite{frechet1948elements} is an extension of the Euclidean mean, widely used to aggregate representations in neural networks such as attention \cite{vaswani2017attention} and batch normalization \cite{ioffe2015batch}, showing strong theoretical and practical interest in Riemannian data analysis.
% The Fr\'echet mean can be expanded with weights (wFM), where the weight $w$ respect the convexity constraint.
Given $\{P_i\}_{i=1}^N$ samples on $\mathcal{S}^+_d$, the Fr\'echet mean is defined as
\begin{align}
    \label{eq6}
    \mu = arg\min_{G \in \mathcal{S}^+_d} \sum^N_{i=1}\delta_R^2(G, P_i).
\end{align}
where $\delta_R$ is the Riemannian distance between two points on the manifold.
% with $w_i \geq 0,\;\sum_{i\leq N}w_i = 1$.
Unfortunately, the solution to the minimization problem in Eq. (\ref{eq6}) is unknown in closed-form, but is usually computed using an iterative solver, \eg, Karcher flow \cite{karcher1977riemannian}, that approximates the solution by computing in tangent space and mapping back to manifold space via logarithmic and exponential maps, respectively. %However, those requires a large computational cost in the approximation.
However, the Fr\'echet mean in Cholesky space is closed and easy to compute.
That is, an iterative operation is not required. 
We call this the \textit{log-Cholesky} mean $\mu_{\mathcal{L}_+}$ and it is defined as \cite{lin2019riemannian}
\begin{align}
    \label{eq7}
    \mu_{\mathcal{L}_+} = \frac{1}{N}\sum^N_{i=1}\lfloor X_i \rfloor + \exp\Big\{N^{-1}\sum^N_{i=1}\log \mathcal{D}(X_i) \Big\}.
\end{align}

\subsection{Manifold ODE} 
% We introduce the ordinary differential equation defined in manifold space \cite{hairer2011ode}. A manifold ODE defines a state $\mathbf{z}:[t_s, t_e]\rightarrow \mathcal{M}$ relating to vector field $f$ in regard to the dynamics between two time points of $t_s$ and $t_e$ with an initial condition $z_s=\mathbf{z}(t_s)$: 
% \begin{align}
%     \label{eq8}
%     \frac{d\mathbf{z}(t)}{dt} = f(\mathbf{z}(t), t) \in \mathcal{T}_{\mathbf{z}(t)}\mathcal{M}.
% \end{align}

We introduce the ordinary differential equations in manifold space. Suppose that $\mathbf{z}(t)$ is a differentiable curve with values in $\mathcal{M}$. Then, by the definition of the tangent space, its derivative $d\mathbf{z}(t)/dt \in \mathcal{T}_{\mathbf{z}(t)}\mathcal{M}\;\forall t$. This assumption leads to the following definition:

\vspace{5pt}\noindent\textbf{Definition 1.} \textit{Let $\mathcal{M}$ be a manifold space. A vector field $f$ in regard to the dynamics is a $\mathcal{C}^{1}$-mapping such that
\begin{align}
    \frac{d\mathbf{z}(t)}{dt} = f(\mathbf{z}(t), t) \in \mathcal{T}_{\mathbf{z}(t)}\mathcal{M}.
\end{align}
A vector field $f$ is called a differential equation on the manifold, and a function $z$ is called an integral curve or simply a solution of the equation.}
\\

\noindent According to Definition 1, a vector field is defined at arbitrary intervals in the manifold space.
We consider the interval $[t_1, t_2]$ such that $z_1 = \mathbf{z}(t_1)$. Since the solution of a differential equation depends on the initial point, we obtain the notation $\varphi_{t_2}(z_1) = \mathbf{z}(t_2)$ for the solution at time $t_2$ with an initial condition $z_1 = \mathbf{z}(t_1)$.

% Here, the curve $\mathbf{z} : [t_s, t_e] \rightarrow \mathcal{M}$ is related to vector field $f$ by the Equation 8.
% Also, local solutions exist under sufficient conditions on $f$ \cite{hairer2011ode}.

\section{Continuous Manifold Learning on Riemannian Cholesky Space}\label{sec:Method}

This section introduces our proposed method of continuous manifold learning in Cholesky space from sequential manifold-valued data for time-series modeling. We illustrate the overall procedures in Fig. \ref{fig_framework}.

\subsection{Mapping to the Riemannian Cholesky Space}
For manifold learning, we first transform the input time-series data into manifold space. 
Specifically, we exploit second-order statistics (\ie, covariance) of features represented via a function $h_\theta$, then use a shrinkage estimator \cite{chen2010shrinkage}, to estimate second-order statistics and thus, represent an SPD matrix in manifold space.
% Specifically, we exploit a second-order statistics (\ie, covariance) of features, estimated by a shrinkage estimator \cite{chen2010shrinkage}, to represent an SPD matrix in manifold space.
% Based on the SPD representation setting, the covariance matrix of features can be a point {\HI in} manifold space. Each point is represented by an SPD matrix, which is estimated by second-order statistics with a shrinkage estimator \cite{chen2010shrinkage}.
Then, an SPD matrix $S$ is decomposed into a lower triangular matrix $X$ and its transpose $X^{\top}$ by Cholesky decomposition $C(S) = XX^\top$.
We express the lower triangular matrix $X$ as the sum of the strictly lower part and the diagonal part as $X = \lfloor X \rfloor + \mathcal{D}(X)$. 

Generally, it is challenging to train a network in SPD Riemannian manifold due to: (1) there are mathematical constraints in optimization, \eg, the convexity of weights in FM estimation \cite{chakraborty2018statistical}, the validity of the output to be an SPD matrix \cite{huang2017riemannian}, and (2) numerical errors that occur by distorting the geometric structure during the learning process, especially when the dimension of a matrix increases \cite{dong2017deep}.
% Through the following mapping, we can achieve several advantages. 
% It is difficult to satisfy the SPD constraint while training a neural network. 
% In particular, as the number of matrix dimensions increase, numerical errors occur during the learning process.
% On the other hand, 
Meanwhile, when conducting operations in Cholesky space, due to the nice properties of the Cholesky matrices, we only consider the constraint that the elements of the diagonal matrix $\mathcal{D}(X)$ should be positive, and the elements in $\lfloor X \rfloor$ are unconstrained. Thus, by taking advantage of operating in the Cholesky space, it becomes relatively easier to solve an optimization problem.
% we only need to satisfy the unconstrained part $\lfloor \cdot \rfloor$ and the positive diagonal part $\mathcal{D}(\cdot)$. In other words, the optimization problem, one of the challenging problems in manifold learning, can be easily handled. Also, we can reduce computation costs in operations in manifold space.

\subsection{Recurrent Network Model in Cholesky Space}
An RNN equipped with gated recurrent units (GRU) \cite{cho2014learning} is successfully used for time-series modeling in various applications. Here, we introduce a novel Riemannian manifold GRU (RGRU) by reformulating gating operations in the Cholesky space.
It is worth noting that the reformulated operations apply not only to GRU but also to LSTM.
% -based model (RGRU) in Cholesky space to preserve the property of lower triangular matrices.
Compared with existing deep learning methods for SPD matrix modeling \cite{chakraborty2018statistical, zhen2021flow, zhen2019dilated, chakraborty2020manifoldnet, huang2017riemannian,brooks2019riemannian, gao2020learning, suh2021riemannian}, as we formulate the optimization problem in a Cholesky space with a unique constraint of positivity in values of the diagonal part of $\mathcal{D}(X)$, the optimization is solved relatively easily and with efficient computations, thus making their computation costs cheap.

% {\cmt In particular, we expect to map a lower triangular matrix into another one while considering time information.[?]} 
Let us reformulate the gating operations and feature representations in terms of the Cholesky geometry with (\romannumeral 1) weighted Fr\'echet mean (wFM), (\romannumeral 2) bias addition, and (\romannumeral 3) non-linearity.
First, the wFM is defined using Eq. (\ref{eq7}) for the weighted combination of Cholesky matrices as follows
%, but by imposing a soft constraint (\ie, positivity) on the diagonal part.
% \begin{align}
%     \operatorname{wFM}&(\left\{X_i\right\}_{i=1,\cdots, N}, \mathbf{w}\in\mathcal{R}_{\geq0})  \nonumber \\
%     =& \frac{1}{N}\sum^N_{i=1}(w_i \cdot \lfloor X_i \rfloor)\nonumber\\
%     &+ \exp\Big\{N^{-1}\sum^N_{i=1}w_i\cdot\log \mathcal{D}(X_i) \Big\}.
% \end{align}
\begin{align}
    \operatorname{wFM}(&\left\{X_i\right\}_{i=1,\cdots, N}, \mathbf{w}\in\mathbb{R}^{N}_{\geq0})  \nonumber\\
    =& \frac{1}{N}\sum^N_{i=1}(w_i \cdot \lfloor X_i \rfloor)
    + \exp\Big\{N^{-1}\sum^N_{i=1}w_i\cdot\log \mathcal{D}(X_i) \Big\}
\end{align}
where $\mathbb{R}_{\geq0}$ denotes a space of non-negative real numbers and $w_i$ denotes the $i$-th element of the weight vector $\mathbf{w}$.
Second, for the bias addition, we define an operator $\oplus$ on $\mathcal{L}_+$ \cite{lin2019riemannian} by 
\begin{align}
    \label{eq9}
    X \oplus Y = \lfloor X\rfloor + \lfloor Y\rfloor + \mathcal{D}(X)\mathcal{D}(Y).
\end{align}
Such an operation $\oplus$ is a smooth commutative group operation on the manifold $\mathcal{L}_+$.  Since this group operation is a bi-invariant metric, we regard the operation for \emph{translation}, analogous to the case in Euclidean space.
Third, regarding the non-linearity, since the off-diagonal values are unconstrained but the diagonal values should be positive, we handle elements of the off-diagonal and the diagonal parts separately by applying two independent activation functions accordingly and then adding the resulting values to obtain a valid Cholesky matrix, \eg, $\text{tanh}(\lfloor X \rfloor) + \text{softplus}(\mathcal{D}(X))$. %it is allowed to use any non-linear function but for the diagonal parta non-linearity function with a range of $[0, \infty)$ was used.

Given a sequence of SPD matrices $\mathbf{X} = \{ X_1, X_2, \cdots, X_T\}$ on $\mathcal{L}_+$ obtained from an input time-series sample by passing through the feature extractor $H_{\theta}$ and the covariance estimation module consecutively, we reformulate the gating and representation operations involved in our RGRU as follows
\begin{align}
    \label{eq10}
    \left\{\begin{array}{c l}
    \mathbf{z}_i &= \sigma(\text{wFM}(\left\{ X_i, H_{i-1}\right\}, W_{z}) \oplus B_z), \\ %\quad \text{(update)},\\
    \mathbf{r}_i &= \sigma(\text{wFM}(\left\{X_i, H_{i-1}\right\}, W_{r}) \oplus B_r), \\ %\quad \text{(reset)}, \\ 
    \mathbf{l}_i &= \text{wFM}(\left\{X_i, \mathbf{r}_i\odot H_{i-1}\right\}, W_{l}) \oplus B_l, \\
    \hat{H}_i &= \text{tanh}(\lfloor \mathbf{l}_i \rfloor) + \text{softplus}(\mathcal{D}(\mathbf{l}_i)), \\%\quad \text{(current memory)}, \\
    H_{i} &= (1-\mathbf{z}_i) \odot H_{i-1} + \mathbf{z}_i \odot \hat{H}_i, %\quad \text{(final memory)}
    \end{array}\right.
\end{align}
where $H_i \in \mathcal{L}_+$ is the hidden state in RGRU, $\mathbf{z}_i\in \mathcal{L}_+, \mathbf{r}_i\in \mathcal{L}_+$ and $\hat{H}_i\in \mathcal{L}_+$ are the update gate, the reset gate, and candidate hidden state, respectively;  $\{W_{z}\in\mathcal{L}_{+},W_{r}\in\mathcal{L}_+,W_{l}\in\mathcal{L}_+\}$ are learnable weights and $\{B_z\in \mathcal{L}_+, B_r\in \mathcal{L}_+$, $B_l\in \mathcal{L}_+\}$ are corresponding learnable biases; 
$\sigma(\cdot)$ is a logistic sigmoid function, tanh($\cdot$) is a hyperbolic tangent function, and softplus$(\cdot)$ is a softplus function, and $\odot$ denotes an operator of element-wise multiplication that satisfies the group axiom. It is worth noting that without loss of generality, any activation function with a positive output range, \eg, sigmoid, can be used in the position of a softplus function. %ReLU

%manifold ODE
\subsection{Neural Manifold ODE}
For continuous manifold learning, we further leverage the technique of neural manifold ODEs \cite{lou2020neural} in sequential data modeling and model the dynamics of $f$ by a neural network with its learnable parameters $\Theta$.
Here, we define both the forward and backward pass gradient computations for network parameter learning. Notably, the forward pass is computed in manifold space but the backward pass is defined solely through an ODE in Euclidean space. %, they do not perfectly match each other. 
\subsubsection{Forward Mode}
Forward mode integration is broadly classified into two groups: a projection method \cite{hairer2011ode} and an implicit method \cite{lou2020neural, hairer2011ode, crouch1993numerical}. 
In this work, we employ an implicit method due to its general applicability in any manifold space and its definition using step-based solvers \cite{bielecki2002estimation, crouch1993numerical}.
% The projection methods embed from manifold space to Euclidean space, integrate with {\HI an} Euclidean solver, and project {\HI back} to the manifold. Although the projection method is simple, there is a problem of generalization in which the projection is not defined in some spaces such as the open ball or the upper half-space.
% In constrast, implicit methods are applicable in any manifold space and are often defined using step-based solvers \cite{bielecki2002estimation, crouch1993numerical}.
Particularly, we use a variant of an Euler method solver with an update step $h_{t+\epsilon} = \exp_{H_t}(\epsilon f_{\Theta}(H_t, t))$ for the Riemannian exponential map \cite{bielecki2002estimation}. 
Here, $\epsilon$ is a time step.

\subsubsection{Backward Mode}
Recently, \cite{chen2018neural} and \cite{lou2020neural} proposed an adjoint sensitivity method to effectively compute the gradients and calculate the derivative of a manifold ODE, independently. In our method, we use the adjoint method to compute the derivative of a manifold ODE.

In differential geometry, $D_xf_{\Theta} : \mathcal{T}_x\mathcal{M} \rightarrow \mathcal{T}_x\mathcal{N}$ defines a derivative of a function $f_{\Theta}: \mathcal{M} \rightarrow \mathcal{N}$ mapping between two manifolds.

\vspace{5pt}\noindent\textbf{Theorem 1.} \cite{lou2020neural} \textit{We define a loss function $E: \mathcal{M} \rightarrow \mathbb{R}$. Suppose that there is an embedding of manifold space $\mathcal{M}$ in Euclidean space $\mathbb{R}^d$. Let the adjoint state be $\mathbf{a}(t) = D_{\mathbf{z}(t)}E$, then the adjoint satisfies}
\begin{align}
    \label{eq11}
    \frac{d\mathbf{a}(t)}{dt} = -\mathbf{a}(t)D_\mathbf{z}(t)f_{\Theta}(\mathbf{z}(t), t)
\end{align}
In the implementation, we construct the differential equation
\begin{align}
    \frac{d\widetilde\Log(H_t)}{dt} = D_{\widetilde\Exp(H_t)}\widetilde\Log\Big(f_{\Theta}(\widetilde\Exp(H_t), t)\Big)
\end{align}
% where $H(t) = H_t$. 
Then, $\widetilde\Log(H_t)$ is solved using the numerical integration technique. Finally, we can update $H_t$ to $H_{t+\epsilon}$.
With the adjoint state, we calculate the gradients with respect to the start and end time points, $t_s$, $t_e$ with the initial condition $H_s$, and the weights $\Theta$.

% Neural ODE are a new family of deep neural network models which can parameterize the continuous dynamics of hidden state.
% We modeled similarly to Neural ODE for the dynamics of the hidden state defined in the manifold space.
% Since the distance between the Euclidean space and manifold space is different, we have to map the hidden state to the tangent space.
% That is, we define a hidden state $H_t$ to be the solution on tangent space to an ODE initial-value problem (IVP):
% where $\widetilde{Log}(\cdot)$ is a logarithmic map in Eq. \ref{eq5} and $H(0)$ is a identity matrix $\mathbf{I}$.
% Then, the output of ODE solver is orthogonally projected to manifold space by exponential map Eq. \ref{eq4} 
\begin{algorithm}[!t]
    % \caption{Continuous manifold Learning on $\mathcal{L}_+$}
    \caption{Pseudo algorithm of ODE-RGRU}\label{alg:algorithm}
    \SetKwInOut{Input}{Input}
    \SetKwInOut{Output}{Output}
    \Input{Time-series data $\mathbf{S}=\{S_i\}_{i=1,\ldots,T}$ and time stamps $\mathbf{t}=\{t_i\}_{i=1,\ldots,T}$, learnable model parameters $\theta,\;\Theta$}
    \Output{$\{H_i\}_{i=1, \cdots, T}$}
    \begin{algorithmic}[1]
    \STATE $ \{S^\prime_i\}_{i=1,\ldots,T} = h_\theta(\mathbf{S})$
    \STATE $\{C^\prime_i\in \mathcal{S}^+_d\}_{i=1,\ldots,T} = \operatorname{SE}( \{S^\prime_i\}_{i=1,\ldots,T})$ \quad \tcp{SE: Shrinkage Estimation}
    \STATE $\{X_{i}X_{i}^\top\}_{i=1,\ldots,T} = \operatorname{CD}(\{C^\prime_i\}_{i=1,\ldots,T})$ \quad \tcp{CD: Cholesky Decomposition}
    \STATE $H_0 = \mathbf{I}$ \quad \tcp{Identity matrix}
    % \WHILE{i $\leq$ N}
    \FOR {$i$ in $1, 2, \cdots, T$}
    \STATE $H^\prime_i = $ ODESolve$(f_\Theta, \widetilde{\Log}(H_{i-1}), (t_{i-1}, t_i))$ 
    %\tcp{Solving ODE in tangent space}
    \STATE $H_i = $ RGRU$(\widetilde{\Exp}(H^\prime_i), X_i)$ \quad \tcp{Eq. (\ref{eq10})}
    % \tcp*[l]{Orthogonal projection and update hidden state}
    \ENDFOR
    \STATE \textbf{return} $\{H_i\}_{i=1, \cdots, T}$
    \end{algorithmic}
\end{algorithm}

\subsection{ODE-RGRU}
Here, we describe the systematic integration of manifold ODE and RGRU in a unified framework for sequential continuous manifold modeling, called ODE-RGRU.
Formally, the ODE-RGRU is formulated as
\begin{align}
    H^\prime_i &= \text{ODESolve}(f_\Theta, \widetilde{\Log}(H_{i-1}), (t_{i-1}, t_i))
    \\H_i &= \text{RGRU}(\widetilde{\Exp}(H^\prime_i), X_i),
\end{align}
where $X_i \in \mathcal{L}_+$ is an input manifold-valued feature and $H_i$ is a latent state at the $i-$th timepoint, for $i = 1, \cdots, T$.
We define such a hidden state $H^{\prime}_i$ to be the solution on tangent space after mapping via $\widetilde{\Log}$ to an ODESolver and then the output of the ODESolver is orthogonally projected to a manifold space by an exponential map $\widetilde{\Exp}$.
The initial hidden state $H_0$ is set to identity matrices $\mathbf{I}$ that satisfy the SPD constraint.
For each observation over time, the corresponding hidden state is updated in the Cholesky space using Eq. (\ref{eq10}), \ie, RGRU.
The overall algorithm of ODE-RGRU is summarized in Algorithm \ref{alg:algorithm}. 
Note that Algorithm \ref{alg:algorithm} focused on the representation of the hidden state, while the proposed model concentrated on mapping to the Cholesky space within the SPD matrix. However, considering the diffeomorphic properties between these two spaces, it is also possible to apply a mapping from the Cholesky space to the SPD space. Thus, depending on the desired objective, mapping between the Cholesky space and the SPD space is feasible, offering flexibility in achieving the intended goals.

\subsection{Computational Time Efficiency}
It should be noted that the computation of manifold space point representations in our proposed model relies on CPU-based operations, which may result in slower training times compared to conventional time series modeling approaches. However, this limitation is mitigated by the inherent statistical characteristics of data and the model's capacity for effective generalization. Consequently, despite the potential trade-off in computational speed, the model enables efficient optimization and exhibits rapid convergence due to its ability to leverage the intrinsic statistical properties of data \cite{kim2020pllay}.

To learn geometric information in a neural network through manifold learning, the internal operations should consider and satisfy manifold constraints. 
Specifically, the SPD matrix appears in various domains, but it is difficult to train a neural network because of rigid constraints, \ie, symmetries and all positive eigenvalues.
To tackle such challenges, \cite{huang2017riemannian} train their model in a compact Stiefel manifold space and \cite{gao2020learning} applied Euclidean operations in a tangent space with a logarithmic map and switched to manifold space with an exponential map.
Most operations are performed through the Eigenvalue decomposition with a computational complexity of $O(k \cdot n^3)$ on $k$ iterations for Fr\'echet mean estimation. That is, as the dimensions of the matrices of interest increase, the computational cost increases exponentially.
Therefore, the previous work \cite{lou2020differentiating} derives explicit gradient expressions so that manifold learning could be performed in a reasonable computational time, but it was limited to hyperbolic space. Consequently, it remains that a computational problem exists in the Riemannian manifold for the Fr\'echet mean estimation, thus making the model's training slow and unstable.
% In this regard, {\cmt the previous work \cite{lou2020differentiating} considered a small-dimensional space} so that manifold learning could be performed in a reasonable computational time. However, it still remains that no closed-form solution exists for the Fr\'echet mean estimation, thus making a model's training unstable.

In this work, we propose to exploit the Cholesky decomposition, which finds a unique lower triangular matrix for an SPD matrix. The nice properties of a Cholesky matrix allow efficient numerical calculations.
%% {\cmt The \textit{\cmt log-Cholesky} metric has the advantage of being affine-invariant, satisfies Lie group bi-invariance, and allows to find a closed-form solution for the wFM of SPD matrices.} 
Moreover, it has simple and easy-to-compute expressions for Riemannian exponential and logarithmic maps. Notably, it provides a closed-form solution for the weighted Fr\'echet mean with a \emph{linear} computational complexity.

\section{Experiments}\label{sec:experiments}
In this section, we evaluate the proposed method for real-world applications of video action recognition, sleep staging classification, and various multivariate time-series modeling tasks by using publicly available datasets.
%% Our method was implemented with PyTorch and trained on GPU NVIDIA GeForce RTX 2080 TI.
Our implementation code used for the experiments is available at \href{https://github.com/ku-milab/Deep-Efficient-Continuous-Manifold-Learning}{"https://github.com/ku-milab/Deep-Efficient-Continuous-Manifold-Learning"} and all the implementation details are provided in our \href{https://github.com/ku-milab/Deep-Efficient-Continuous-Manifold-Learning}{GitHub repository}.

\begin{table}[t]
    \centering
    \caption{Performance (mean$\pm$std, \%) on an action recognition task (UCF11).} %{\sw \# Params. denotes the number of learnable parameters.}}
    \label{table3}
    \begin{tabular}{c|cc}
        \toprule
        Method                                                                       & \# Parameters   & Accuracy                  \\ \midrule\midrule 
        ODE-RGRU (Ours)                                                          & 66,767     & \textbf{89.4 $\pm$ 1.2} \\
        \begin{tabular}[c]{@{}c@{}}ManifoldDCNN \cite{zhen2019dilated}\end{tabular}      & 3,658       & 82.3 $\pm$ 1.8              \\
        \begin{tabular}[c]{@{}c@{}}SPDSRU \cite{chakraborty2018statistical}\end{tabular} & \textbf{3,445}       & 78.4 $\pm$ 1.4              \\ \midrule
        \begin{tabular}[c]{@{}c@{}}TT-GRU \cite{yang2017tensor}\end{tabular}             & 202,667    & 81.3 $\pm$ 1.1              \\
        \begin{tabular}[c]{@{}c@{}}TT-LSTM \cite{yang2017tensor} \end{tabular} &  6,176     &79.6 $\pm$ 3.5              \\
        GRU                                                                          & 18,602,606 & 79.3 $\pm$ 3.8              \\
        LSTM                                                                         & 24,791,606 & 71.4 $\pm$ 2.7              \\ \bottomrule
    \end{tabular}
\end{table}

\subsection{Datasets and Preprocessing}
%% {\HI}In order for fair comparison, we used two publicly available datasets.}
\subsubsection{Action Recognition}
% \vspace{5pt}\noindent\textbf{Action Recognition.}
For the action recognition task, we used the UCF11 \cite{liu2009recognizing} dataset.
The UCF11 dataset is composed of 1,600 video clips, with 11 action categories, \eg, basketball shooting, biking/cycling, diving, golf swinging, \etc. The length of the video frames varies from 204 to 1492 per clip with a resolution of each frame being $320 \times 240$. Similar to previous work \cite{zhen2019dilated, chakraborty2018statistical}, we downsized all frames to $160 \times 120$ and sampled 50 frames from each clip, \ie, $N=50$, with an equal time gap between frames. Then, we performed the $5$-fold cross-validation for fair evaluation, while keeping the class balance.

\subsubsection{Sleep Staging Classification}
% \vspace{5pt}\noindent\textbf{Sleep Staging Classification. }
We used the SleepEDF-20 \cite{goldberger2000physiobank} dataset for the sleep staging classification task. The SleepEDF-20 is the Sleep Cassette that comprises 20 subjects (10 males and 10 females) aged 25-34. Two polysomnographies were recorded during day-night periods, except for subject 13 who lost the second night due to a device problem. All signals were segmented into lengths of 30 seconds and manually labeled into 8 categories, `wake,' `rapid eye movement (REM),' `non REM1/2/3/4,'  `movement,' and `unknown,' by sleep experts  \cite{rechtschaffen1968manual}. In this work, similar to the previous studies \cite{phan2018automatic, phan2018joint, supratak2017deepsleepnet, phan2021xsleepnet}, we merged `non REM3' and `non REM4' stages into a `non REM3' stage and removed samples labeled as `movement' and `unknown.' Here, we used spectrogram data rather than raw signals \cite{phan2019seqsleepnet}. To achieve this, we sampled all signals at 100Hz rate and applied short-time Fourier transformation to compute spectrograms. For this experiment, we used single-channel EEG and 2-channel EEG/EOG samples. Note that we did not exploit electromyogram recordings due to their incompleteness. As a result, we obtained a multi-channel signal $S \in \mathbb{R}^{F \times T \times C}$ where $F$, $T$, and $C$ denote the number of frequency bins, the number of spectral columns, and the number of channels, respectively. We set $F=129$, $T=29$, $C=1$ or $C=2$ (for the single-channel EEG or two-channel EEG/EOG). For the staging classification, $10$ continuous segments of data were used as input, \ie, $N=10$.  For a fair evaluation, we conducted a $20$-fold cross-validation by maintaining a class balance similar to \cite{phan2018automatic, phan2018joint, supratak2017deepsleepnet, phan2021xsleepnet}.

% \vspace{5pt}\noindent\textbf{Multivariate Time Series Classification. }
\subsubsection{Multivariate Time Series Classification}
To evaluate various sensor data in the classification task not only for sleep staging classification, we used the UEA archive \cite{bagnall2018uea}, a well-known multivariate time-series data. The UEA archive consists of 30 real-world multivariate time-series data for various classification applications, \eg, Human Action Recognition, Motion classification, \etc. The dimension of the archive range from 2 dimensions in AtrialFibrillation to 1345 in DuckDuckGeese, and the time length is from 8 in PenDigits to 17984 in EigenWorms. The dataset is divided into train and test sets with the train size ranging from 12 to 30000. Detailed information about each dataset can be found in \cite{bagnall2018uea}.

\subsubsection{Time Series Forecasting \& Imputation}

For the forecasting task, we used four public datasets, including three ETT datasets \cite{zhou2021informer} and the Electricity dataset \cite{asuncion2007uci}. The ETT is the dataset that collects electricity transformer temperature, which is an important indicator in power long-term deployment. The ETT dataset consists of ETTh$_1$, ETTh$_2$ at a 1-hour level and ETTm$_1$ at a 15-minute level, and each point is expressed as an oil temperature and 6 power load features. Electricity dataset is the measurement of electric power consumption that contains 2,075,259 measurements.

For the imputation task, we collected data from the TADPOLE database\footnote{https://tadpole.grand-challenge.org/Data}, which is based on the Alzheimer's Disease Neuroimaging Initiative (ADNI) cohort. This dataset comprises information from 1,737 patients and encompasses 1,500 biomarkers obtained over 12,741 visits spanning 22 time periods. Although the TADPOLE database offers numerous biomarkers for predicting the Alzheimer's disease (AD) spectrum, our study focused on six volumetric MRI features, T1-weighted MRI scans, and cognitive test scores, such as the mini-mental state exam (MMSE), Alzheimer's disease assessment scale (ADAS)-cog11, and ADAS-cog13.

\begin{table*}[th]
    \centering
    \caption{Performance on sleep staging classification task (SleepEDF-20). We computed accuracy (Acc), Cohen's kappa value ($\kappa$), macro F1-score (MF1), averaged sensitivity (Sen), and averaged specificity (Spec).}
    \label{table4}
    \begin{tabular}{c|ccccc|ccccc}
    \toprule
        \multirow{2}{*}{Method} & \multicolumn{5}{c|}{EEG} & \multicolumn{5}{c}{EEG/EOG} \\ \cline{2-11} 
         & Acc & $\kappa$ & MF1 & Sen & Spec & Acc & $\kappa$ & MF1 & Sen & Spec \\ \midrule\midrule
        ODE-RGRU (Ours) & \textbf{86.3} & 0.807 & 80.2 & \textbf{80.6} & 96.3 & \textbf{86.6} & 0.812 & 80.3 & \textbf{81.4} & \textbf{96.4} \\ 
        ODE-RGRU (w/o ODEFunc.) (Ours) &84.1&0.774&74&75&95.6& 84.3 & 0.777 & 74.8 & 75.7 & 95.7 \\
        XSleepNet  \cite{phan2021xsleepnet}& \textbf{86.3} & \textbf{0.813} & \textbf{80.6} & 80.2 & \textbf{96.4} & 86.4 & \textbf{0.813} & \textbf{80.9} & 79.9 & 96.2 \\ 
        SeqSleepNet \cite{phan2019seqsleepnet}& 85.2 & 0.798 & 78.4 & 78.0 & 96.1 & 86.0 & 0.809 & 79.7 & 79.2 & 96.2 \\
        FCNN + RNN  \cite{phan2021xsleepnet} & 81.8 & 0.754 & 75.6 & 75.7 & 95.3 & 83.5 & 0.775 & 77.7 & 77.2 & 95.5 \\
        DeepSleepNet \cite{supratak2017deepsleepnet} & - & - & - & - & - & 82.0 & 0.760 & 76.9 & - & - \\
        U-time \cite{perslev2019u}& - & - & 79.0 & - & - & - & - & - & - & - \\
        IITNet \cite{seo2020intra}& 83.9 & 0.780 & 77.6 & - & - & - & - & - & - & - \\ 
        \bottomrule
        \end{tabular}
\end{table*}

\begin{table*}[th]
\centering
\caption{Performance on UEA archive. We compute the accuracy of all comparison methods, total best accuracy, average rank, and Wilcoxon-signed rank test.}
\label{table5}
\scalebox{.9}{
\begin{tabular}{c|ccccccccccc}
\toprule
\multirow{2}{*}{Dataset} & EDI  & DTWI  & DTWD  & MLSTM & WEASEL & NS  & TapNet & ShapeNet & TST & TS2Vec & ODE-RGRU  \\
 & \cite{bagnall2018uea} & \cite{bagnall2018uea} & \cite{bagnall2018uea} & -FCNs~\cite{karim2019multivariate} & +MUSE~\cite{schafer2017multivariate} & \cite{franceschi2019unsupervised} & \cite{zhang2020tapnet} & \cite{li2021shapenet} & \cite{zerveas2021transformer} & \cite{yue2022ts2vec} & (Ours) \\

\midrule\midrule
ArticularyWordRecognition & 0.970 & 0.980 & 0.987 & 0.973 & \textbf{0.990} & 0.987 & 0.987 & 0.987 & 0.977 & 0.987 & 0.973 \\
AtrialFibrillation        & 0.267 & 0.267 & 0.220 & 0.267 & 0.333 & 0.133 & 0.333 & \textbf{0.400} & 0.067 & 0.200 & 0.333 \\
BasicMotions              & 0.676 & \textbf{1.000} & 0.975 & 0.950 & \textbf{1.000} & \textbf{1.000} & \textbf{1.000} & \textbf{1.000} & 0.975 & 0.975 & \textbf{1.000} \\
CharacterTrajectories     & 0.964 & 0.969 & 0.989 & 0.985 & 0.990 & 0.994 & \textbf{0.997} & 0.980 & 0.975 & 0.995 & 0.991 \\
Cricket                   & 0.944 & 0.986 & \textbf{1.000} & 0.917 & \textbf{1.000} & 0.986 & 0.958 & 0.986 & \textbf{1.000} & 0.972 & 0.986  \\
DuckDuckGeese             & 0.275 & 0.550 & 0.600 & 0.675 & 0.575 & 0.675 & 0.575 & \textbf{0.725} & 0.620 & 0.680 & 0.700 \\
EigenWorms                & 0.549 & -     & 0.618 & 0.504 & \textbf{0.890} & 0.878 & 0.489 & 0.878 & 0.748 & 0.847 & 0.847 \\
Epilepsy                  & 0.666 & 0.978 & 0.964 & 0.761 & \textbf{1.000} & 0.957 & 0.971 & 0.987 & 0.949 & 0.964 & 0.964 \\ 
ERing                     & 0.133 & 0.133 & 0.133 & 0.133 & 0.133 & 0.133 & 0.133 & 0.133 & \textbf{0.874} & \textbf{0.874} & 0.748 \\ 
EthanolConcentration      & 0.293 & 0.304 & 0.323 & 0.373 & \textbf{0.430} & 0.236 & 0.323 & 0.312 & 0.262 & 0.308 & 0.373 \\ 
FaceDetection             & 0.519 & -     & 0.529 & 0.545 & 0.545 & 0.528 & 0.556 & 0.602 & 0.534 & 0.501 & \textbf{0.620} \\ 
FingerMovements           & 0.550 & 0.520 & 0.530  & 0.580  & 0.490  & 0.540  & 0.530  & 0.580 & 0.560 & 0.480 & \textbf{0.600} \\ 
HandMovementDirection     & 0.278 & 0.306 & 0.231 & 0.365 & 0.365 & 0.270 & \textbf{0.378} & 0.338 & 0.243 & 0.338 & 0.311 \\ 
Handwriting               & 0.200 & 0.316 & 0.286 & 0.286 & \textbf{0.605} & 0.533 & 0.357 & 0.451 & 0.225 & 0.515 & 0.457 \\ 
Heartbeat                 & 0.619 & 0.658 & 0.717 & 0.663 & 0.727 & 0.737 & 0.751 & \textbf{0.756} & 0.746 & 0.683 & 0.751 \\
InsectWingbeat            & 0.128 & -     & -     & 0.167 & -     & 0.160 & 0.208 & 0.250 & 0.105 & 0.466 & \textbf{0.548} \\ 
JapaneseVowels            & 0.924 & 0.959 & 0.949 & 0.976 & 0.973 & 0.989 & 0.965 & 0.984 & 0.978 & 0.984 & \textbf{0.992} \\
Libras                    & 0.833 & \textbf{0.894} & 0.870 & 0.856 & 0.878 & 0.867 & 0.850 & 0.856 & 0.656 & 0.867 & 0.861 \\ 
LSST                      & 0.456 & 0.575 & 0.551 & 0.373 & 0.590 & 0.558 & 0.568 & 0.590 & 0.408 & 0.537 & \textbf{0.595} \\ 
MotorImagery              & 0.510 & -     & 0.500 & 0.510 & 0.500 & 0.540 & 0.590 & 0.610 & 0.500 & 0.510 & \textbf{0.630} \\
NATOPS                    & 0.850 & 0.850 & 0.883 & 0.889 & 0.870 & 0.944 & 0.939 & 0.883 & 0.850 & 0.928 & \textbf{0.950} \\ 
PEMS-SF                   & 0.705 & 0.734 & 0.711 & 0.699 & -     & 0.688 & 0.751 & 0.751 & 0.740 & 0.682 & \textbf{0.763} \\
PenDigits                 & 0.973 & 0.939 & 0.977 & 0.978 & 0.948 & 0.983 & 0.980 & 0.977 & 0.560 & \textbf{0.989} & 0.986 \\
Phoneme                   & 0.104 & 0.151 & 0.151 & 0.110 & 0.190 & 0.246 & 0.175 & 0.298 & 0.085 & 0.233 & \textbf{0.302} \\ 
RacketSports              & 0.868 & 0.842 & 0.803 & 0.803 & \textbf{0.934} & 0.862 & 0.868 & 0.882 & 0.809 & 0.855 & 0.868 \\
SelfRegulationSCP1        & 0.771 & 0.765 & 0.775 & \textbf{0.874} & 0.710 & 0.846 & 0.652 & 0.782 & 0.754 & 0.812 & 0.823 \\ 
SelfRegulationSCP2        & 0.483 & 0.533 & 0.539 & 0.472 & 0.460 & 0.556 & 0.550 & \textbf{0.578} & 0.550 & \textbf{0.578} & 0.550 \\ 
SpokenArabicDigits        & 0.967 & 0.959 & 0.963 & \textbf{0.990} & 0.982 & 0.956 & 0.983 & 0.975 & 0.923 & 0.988 & 0.977 \\ 
StandWalkJump             & 0.200 & 0.333 & 0.200 & 0.067 & 0.333 & 0.400 & 0.400 & \textbf{0.533} & 0.267 & 0.467 & 0.400 \\
UWaveGestureLibrary       & 0.881 & 0.868 & 0.903 & 0.891 & \textbf{0.916} & 0.884 & 0.894 & 0.906 & 0.575 & 0.906 & 0.834 \\
\midrule
Total Best acc & 0 & 2 & 1 & 2 & 9 & 1 & 3 & 6 & 2 & 3 & \textbf{10} \\
Average Rank & 6.8 & 6.0 & 5.4 & 5.2 & 4.2 & 4.2 & 4.0 & 2.9 & 6.2 & 3.9 & 2.6 \\
Wilcoxon Test p-value & 0.000 & 0.000 & 0.000 & 0.000 & 0.155 & 0.013 & 0.003 & 0.802 & 0.000 & 0.094 & - \\
\bottomrule
\end{tabular}}
\end{table*}

\begin{table}[t]
    \centering
    \caption{Performance on irregular CharacterTrajectories. We computed the mean and variance of accuracy on five repeats.}
    \label{table6}
    \begin{tabular}{c|ccc}
    \toprule
        \multirow{2}{*}{Method} & \multicolumn{3}{c}{Test Accuracy}\\ \cline{2-4} 
         &30\% dropped&50\% dropped&70\% dropped\\ \midrule\midrule
        ODE-RGRU (Ours) &\textbf{98.9 $\pm$ 0.2} &98.3 $\pm$ 0.2&\textbf{98.6 $\pm$ 0.3}\\ 
        Neural CDE \cite{kidger2020neural}&98.7 $\pm$ 0.8&\textbf{98.8 $\pm$ 0.2}&98.6 $\pm$ 0.4\\ 
        ODE-RNN \cite{rubanova2019latent}&95.4 $\pm$ 0.6&96 $\pm$ 0.3&95.3 $\pm$ 0.6\\
        GRU-D \cite{che2018recurrent}&94.2 $\pm$ 2.1&90.2 $\pm$ 4.8&91.9 $\pm$ 1.7\\
        GRU-$\Delta$t \cite{kidger2020neural}&93.6 $\pm$ 2&91.3 $\pm$ 2.1&90.4 $\pm$ 0.8\\ 
        \bottomrule
    \end{tabular}
\end{table}

\subsection{Experimental Setting}
\subsubsection{Action Recognition}
In our work, we compared the proposed method with state-of-the-art manifold learning methods and typical deep learning models in the action recognition task. For the baseline methods representing data in Euclidean space, we exploited GRU \cite{cho2014learning}, LSTM \cite{hochreiter1997long}, TT-GRU, and TT-LSTM \cite{yang2017tensor}. For GRU and LSTM, we first extracted spatial features using 3 convolutional layers with a kernel size of $7\times 7$ and channel dimensions of 10, 15, and 25, respectively. Further, batch normalization \cite{ioffe2015batch}, leaky rectified linear units (LeakyReLU) \cite{maas2013rectifier}, and max pooling were followed after each convolutional layer. Then, the extracted spatial features were flattened and fed into the recurrent models, \ie, GRU and LSTM, both of which had $50$ hidden nodes. For TT-GRU and TT-LSTM, we reshaped an input sample to $8 \times 20 \times 20 \times 18$ and set the rank size to $1\times 4\times 4\times 4\times 1$. Finally, we obtained the $4 \times 4 \times 4\times 4$ output shape.

For the other baseline methods considered in the action recognition experiments, we employed SPDSRU \cite{chakraborty2018statistical} and ManifoldDCNN \cite{zhen2019dilated}. For covariance matrices used in SPDSRU and ManifoldDCNN, we adopted a covariance block analogous to \cite{yu2017second} to get the reported testing accuracies. Then, for SPDSRU, 2 convolutional layers with $7\times7$ kernels, $7$ channels for the final output were chosen, therefore the dimension of the covariance matrix was $8\times 8$ the same as in \cite{chakraborty2018statistical}. Finally, based on \cite{zhen2019dilated}, ManifoldDCNN also used 2 convolutional layers with $7\times7$ kernels and set $6$ channels for the final output, \ie, the dimension covariance matrix was $7\times7$.

While training our proposed method, we chose 3 convolutional layers with a kernel size of $7 \times 7$ and exploited batch normalization \cite{ioffe2015batch}, LeakyReLU activation \cite{maas2013rectifier}, and max pooling for $h_\theta$. Further, we set the number of output dimensions to $32$, \ie, the dimension of the SPD matrix was $32\times 32$. Therefore, the diagonal matrix and strictly lower triangular matrix through Cholesky decomposition were expressed as a 32-dimensional vector and a vector of $32\times (32-1)/2$ dimension, respectively. Moreover, similar to neural ODEs \cite{chen2018neural}, we employed a multilayer perceptron (MLP) with a $\tanh$ activation of $f_\Theta$. Finally, we set the number of hidden units to $100$ for RGRU.

\subsubsection{Sleep Staging Classification}
%sleep staging classification
For the sleep staging classification experiments, we compared our proposed method with the state-of-the-art methods of XSleepNet \cite{phan2021xsleepnet}, SeqSleepNet \cite{phan2019seqsleepnet}, FCNN+RNN \cite{phan2021xsleepnet}, DeepSleepNet \cite{supratak2017deepsleepnet}, U-time \cite{perslev2019u}, and IITNet \cite{seo2020intra}. We report the performance of those competing methods by taking from \cite{phan2021xsleepnet}.

By following \cite{phan2021xsleepnet}, we used a filterbank layer \cite{phan2018dnn} for time-frequency input data. Through the learnable filterbank layer, the spectral dimension of the input data was reduced from $F = 129$ to $32$. Here, 2 convolutional layers followed by batch normalization and a LeakyReLU activation were used for $h_\theta$. The output dimension was set as $8$, \ie, the dimension of the SPD matrix was $8\times8$. We also used an MLP with a $\tanh$ for $f_\Theta$ and set $50$ hidden units for RGRU. Finally, since the sleep staging of a sample is related to neighboring samples, we also exploited a bidirectional structure in RGRU to consider such useful information, the same as in other competing methods.

%UEA dataset
\subsubsection{Multivariate Time Series Classification}
In the UEA archive experiment, we compared our proposed method with eight different methods, including three benchmarks \cite{bagnall2018uea}, a bag-of-pattern based approach \cite{schafer2017multivariate}, and deep learning-based methods \cite{karim2019multivariate, franceschi2019unsupervised, zhang2020tapnet, li2021shapenet, yue2022ts2vec, zerveas2021transformer}. The three benchmarks are Euclidean Distance (EDI), which is a 1-nearest neighbor with distance functions, dimension-independent dynamic time warping (DTWI), and dimension-dependent dynamic time warping (DTWD) \cite{bagnall2018uea}. WEASEL-MUSE \cite{schafer2017multivariate} is a bag-of-pattern-based approach with statistical feature selection. For deep learning-based methods, MLSTM-FCN \cite{karim2019multivariate} is the multivariate extension of the LSTM-FCN framework with a squeeze-and-excite block. Negative samples (NS) \cite{franceschi2019unsupervised} is an unsupervised learning method with triplet loss through several negative samples. TapNet \cite{zhang2020tapnet} is an attentional prototype network with limited training labels. ShapeNet \cite{li2021shapenet} is a shapelet-neural network for multivariate time-series classification tasks. TST \cite{zerveas2021transformer} is a transformer-based unsupervised learning approach. We reported the performance of the baseline results from the original papers \cite{bagnall2018uea, karim2019multivariate, schafer2017multivariate, franceschi2019unsupervised, zhang2020tapnet, li2021shapenet, zerveas2021transformer} and \cite{yue2022ts2vec}, respectively.

% parameters
To train the model, we first divided the data with sliding time windows. Since the time resolution is high, the time window is used for the efficiency of the calculation. Therefore, the input value was divided into 5 time intervals, and the parameter was set from 1 to 30 according to the time length in each dataset. For the encoder part, we chose 3 convolutional layers with a kernel size of $2$ and batch normalization, LeakyReLU activation. We set the number of output dimensions to $32$ and the number of hidden units $32$ for RGRU. All parameters were adjusted according to each dataset's dimension and time length ratio.
%\footnote{See Supplementary A for detailed model architectures used in each experiments.}

\subsubsection{Time Series Forecasting \& Imputation}
For the forecasting experiment, we compared our proposed method with three different methods, including TCN \cite{bai2018empirical}, Informer \cite{zhou2021informer}, and TS2Vec \cite{yue2022ts2vec}. We reported the performance of those competing methods by taking from \cite{yue2022ts2vec}. For the imputation task, a standard LSTM with a mean (LSTM-M), forward (LSTM-F) imputation, MRNN \cite{yoon2018estimating}, and MinimalRNN \cite{nguyen2020predicting} were used as comparison methods. 

To train the model for forecasting tasks, we set up the encoder using experimental settings in UEA. For the decoder part, we chose 1 transposed convolutional layer with LeakyReLU activation and 1 convolutional layer for the prediction part. All parameters were adjusted according to the dimension and time length in each dataset and the prediction length $H$.

To ensure a fair experimental comparison in imputation tasks, we selected 11 out of the 22 available AD-progression prediction time sequences. In order to maintain consistency, subjects who did not have a baseline visit or had fewer than three visits were excluded from our analysis. As a result, our final dataset consisted of 691 subjects. Our proposed method incorporates 2 convolutional layers followed by batch normalization and LeakyReLU activation. The output dimension and RGRU hidden units were set as 32 and 16, respectively.

\subsubsection{Training Details} 
For all experiments except forecasting and imputation, we used a multi-class cross-entropy loss and employed an Adam optimizer with a learning rate of $10^{-4}$ and an $\ell_2$ regularizer with a weighting coefficient of $10^{-3}$. We trained our proposed network for 1,000 (action recognition), 20 (sleep staging classification) epochs, and 400 (UEA archive, ETT, Electricity, and ADNI) iterations with a batch size of 32. For forecasting and imputation, we used a mean squared error (MSE) loss.
For the learnable parameters of RGRU, which should be in Cholesky space, \ie, $W_z, W_r, W_l, B_z, B_r, B_l\in \mathcal{L}_+$, we only needed to care about the diagonal elements. In order for that, we simply applied an absolute operator during backpropagation.

\begin{table*}[!t]
    \centering
    \caption{Performance of time series imputation in terms of MAPE and $R^2$. ($*$: $p<0.05$)}
    \label{table7}
    \begin{tabular}{ccccccc}
    \toprule
    \multirow{2}{*}{Method} & \multicolumn{2}{c}{MMSE} & \multicolumn{2}{c}{ADAS-cog11} & \multicolumn{2}{c}{ADAS-cog13} \\ \cmidrule{2-7}
    & MAPE $\downarrow$ &$R^2$ $\uparrow$ & MAPE $\downarrow$ &$R^2$ $\uparrow$ & MAPE $\downarrow$ & $R^2$ $\uparrow$ \\
    \midrule\midrule
    LSTM-M & 0.173$\pm$0.030$^*$ & -0.412$\pm$1.143$^*$ & 0.929$\pm$0.433$^*$ & 0.321$\pm$0.173$^*$ & 0.863$\pm$0.289$^*$ & 0.302$\pm$0.267$^*$\\
    LSTM-F & 0.235$\pm$0.110$^*$ & -0.053$\pm$0.495$^*$& 0.829$\pm$0.353$^*$ & 0.198$\pm$0.494$^*$& 0.790$\pm$0.152$^*$ & 0.177$\pm$0.468$^*$ \\
    MRNN \cite{yoon2018estimating} & 0.149$\pm$0.031$^*$ & 0.168$\pm$0.284$^*$ &0.930$\pm$0.224$^*$ & 0.262$\pm$0.184$^*$& 0.920$\pm$0.234$^*$ & 0.263$\pm$0.187$^*$\\
    MinimalRNN \cite{nguyen2020predicting} & 0.175$\pm$0.052$^*$ & 0.472$\pm$0.116$^*$ &0.565$\pm$0.142$^*$ & 0.569$\pm$0.038$^*$& 0.451$\pm$0.111& 0.635$\pm$0.049$^*$\\
    Ours & \textbf{0.099$\pm$0.018} & \textbf{0.608$\pm$0.067} & \textbf{0.441$\pm$0.043} & \textbf{0.689$\pm$0.036} & \textbf{0.403$\pm$0.043} & \textbf{0.726$\pm$0.034} \\
    \bottomrule
    \end{tabular}
\end{table*}

\begin{table}[t]
\centering
\caption{Performance on a multivariate time series forecasting task. We computed the mean squared error (MSE) of all comparison methods.}
\label{table8}
\begin{tabular}{cc|cccc}
\toprule
\multirow{2}{*}{Dataset} & \multirow{2}{*}{$H$} & TCN & Informer & TS2Vec & ODE-RGRU  \\
 & & \cite{bai2018empirical} & \cite{zhou2021informer} & \cite{yue2022ts2vec} & (Ours) \\
\midrule\midrule
\multirow{5}{*}{ETTh$_1$} & 24  & 0.767 & \textbf{0.577} & 0.599 & 0.643 \\
                          & 48  & 0.713 & 0.685 & 0.629 & \textbf{0.616} \\
                          & 168 & 0.995 & 0.931 & \textbf{0.755} & 0.838 \\
                          & 336 & 1.175 & 1.128 & 0.907 & \textbf{0.885} \\
                          & 720 & 1.453 & 1.215 & 1.048 & \textbf{0.862} \\
\midrule
\multirow{5}{*}{ETTh$_2$} & 24  & 1.365 & 0.720 & \textbf{0.398} & 0.781 \\
                          & 48  & 1.395 & 1.457 & \textbf{0.580} & 0.881 \\
                          & 168 & 3.166 & 3.489 & \textbf{1.901} & 3.945 \\
                          & 336 & 3.256 & 2.723 & \textbf{2.304} & 2.472 \\
                          & 720 & 3.690 & 3.467 & 2.650 & \textbf{2.283} \\
\midrule
\multirow{5}{*}{ETTm$_1$} & 24  & 0.324 & \textbf{0.323} & 0.443 & 0.672\\
                          & 48  & \textbf{0.477} & 0.494 & 0.582 & 0.783\\
                          & 96  & 0.636 & 0.678 & \textbf{0.622} & 0.648\\
                          & 288 & 1.270 & 1.056 & \textbf{0.709} & 0.765\\
                          & 672 & 1.381 & 1.192 & \textbf{0.786} & 0.835\\
\midrule
\multirow{5}{*}{Electric.} & 24  & 0.305 & 0.312 & 0.287 & \textbf{0.260} \\
                           & 48  & 0.317 & 0.392 & \textbf{0.307} & 0.396 \\
                           & 168 & 0.358 & 0.515 & 0.332 & \textbf{0.328} \\
                           & 336 & 0.349 & 0.759 & 0.349 & \textbf{0.336} \\
                           & 720 & 0.447 & 0.969 & \textbf{0.375} & \textbf{0.375} \\

\midrule
\multicolumn{2}{c}{Total best MSE} & 1 & 1 & \textbf{10} & 8 \\
\bottomrule
\end{tabular}
\end{table}

\subsection{Experimental Results and Analysis}
\subsubsection{Classification Performance}

\vspace{5pt}\noindent\textbf{Action Recognition. } As presented in TABLE \ref{table3}, the proposed method outperformed the competing baseline methods with a large margin (7.1\% $\sim$ 18.0\%) in the action recognition task. Notably, the proposed method achieved the best performance although the number of its tunable parameters was significantly smaller than the other Euclidean representation-based methods. Based on these promising results, we presume that the proposed method could learn geometric features in complex actions efficiently. 

% \subsubsection{Sleep Staging Classification}
\vspace{5pt}\noindent\textbf{Sleep Staging Classification. }We evaluated the methods in terms of accuracy (Acc), Cohen's kappa value ($\kappa$) \cite{mchugh2012interrater}, macro F1-score (MF1) \cite{yang1999re}, averaged sensitivity (Sen), and averaged specificity (Spec) and reported the results in TABLE \ref{table4}. In the single-channel EEG analysis, our proposed method showed an accuracy of 86.3\% and a sensitivity of 80.6\%, which were the same or higher than the comparative methods. 
Although XSleepNet \cite{phan2021xsleepnet} achieved slightly higher performance for Cohen's kappa value, the macro F1-score, and the specificity, the proposed method also showed comparable and plausible performance. Furthermore, it is worth noting that XSleepNet used both raw signals and spectrograms for its decision-making. For the two-channel EEG/EOG case, the proposed method outperformed all baseline methods for metrics of accuracy, sensitivity, and specificity. In contrast, our method slightly underperformed XSleepNet, which exploited complement data types.

% \subsection{UEA archive}
\vspace{5pt}\noindent\textbf{Multivariate Time Series Classification. } The experimental results from the UEA archive are shown in TABLE \ref{table5}. For each dataset, we evaluated the accuracy and computed the total best accuracy, average rank, and Wilcoxon-signed rank test by following \cite{li2021shapenet}. Our proposed method showed the best performance in 10 datasets, and in particular, Ering and InsectWingbeat showed overwhelming performance compared to the competing baseline methods. Also, compared with distance-based methods such as NS \cite{franceschi2019unsupervised}, TapNet \cite{zhang2020tapnet}, and ShapeNet \cite{li2021shapenet}, it showed sufficiently high performance even on small-sized datasets (\eg, AtrialFibrillation, BasicMotions). This shows that our proposed method learns the trajectory of time well to discover a  sufficient representation even with limited training samples. Our method ranked the top on average among the competing methods considered in this work by showing the highest performance over 10 datasets in general time-series classification tasks.
We also conducted a Wilcoxon-signed rank test following in \cite{li2021shapenet, demvsar2006statistical, holm1979simple} for the statistical validity of our proposed method. As presented in the last row of TABLE \ref{table5}, all results are statistically significant, except for WEASEL+MUSE and ShapeNet. 

Compared to WEASEL+MUSE, we observed that ODE-RGRU consistently outperforms, particularly in scenarios involving high-dimensional or large datasets such as InsectWingbeat and PEMS-SF. Conversely, WEASEL+MUSE often grappled with challenges, including memory issues~\cite{zhang2020tapnet}, which hindered its execution. The limitation in WEASEL+MUSE can be attributed to the necessity of generating symbols (words) for each sub-sequence, considering its length and dimension. Notably, in the case of high-dimensional time series data, the size of the symbol dictionary can significantly increase, making the approach less suitable for efficiently handling large datasets.
% {\revision Indeed, despite the similarities in performance, it is important to note that ShapeNet requires the identification of shapelets for each individual dataset, whereas our proposed model allows for representation learning across all datasets using a single model.}

ShapeNet has empirically exhibited performance levels comparable to our proposed method, but we argue it has certain inherent limitations. Specifically, ShapeNet requires the identification of shapelets tailored to each dataset. This procedure can be time-consuming and might not fully encompass crucial details regarding variables and subsequence positions. In contrast, our proposed approach addresses these limitations by acquiring universal representations encompassing essential temporal and geometric information within time series data. This methodological disparity renders our approach versatile and adaptable across various tasks, as it prioritizes the capture of general time series characteristics.

\subsubsection{Forecasting \& Imputation Performance}

We evaluated the competing methods in terms of MSE and reported the results in TABLE~\ref{table8}. Our proposed method demonstrated the best performance in 8 cases and exhibited comparable results in most instances. However, the proposed method generally showed competitive performance in long-term predictions but tended to show somewhat poor performance in short-term predictions.

This discrepancy might be attributed to using a sliding window for conducting prediction tasks on long-time series data. While this approach extracts temporal and geometric information into the manifold space, it introduces challenges in incorporating semantic information into the prediction process. Although TS2Vec~\cite{yue2022ts2vec}, which learned semantic level time-series representation, showed better performance overall, our proposed method showed plausible performance in the ETTh$_1$ and electricity datasets.
% We evaluated the competing methods in terms of MSE and reported the results in TABLE 8. Our proposed method demonstrated the best performance in 8 cases and exhibited comparable results in most instances. {\rev However, it is worth noting that the proposed method generally displayed competitive performance when predicting long-term time points but performed rather poorly when predicting short-term time points.

% This discrepancy can be attributed to using a sliding window when conducting prediction tasks on long-time series data. While this approach effectively extracts temporal and geometric information to the manifold space, it introduces challenges in incorporating semantic information into the prediction process.} Although TS2Vec [7], which learned semantic level time-series representation, showed better performance overall, our proposed method showed plausible performance in the ETTh1 and electricity datasets.

% {\revision We evaluated the competing methods in terms of MSE and reported the results in TABLE \ref{table8}. Our proposed method showed the best performance in 8 cases and showed comparable performance in all cases except for $H=48$ in the electricity dataset. Although TS2Vec \cite{yue2022ts2vec}, which learned semantic level time-series representation, showed better performance overall, our proposed method showed plausible performance in the ETTh1 and electricity datasets.

We also conducted cognitive score prediction to demonstrate the imputation ability of our method. We evaluated the metrics of mean absolute percentage error (MAPE) and coefficient of determination ($R^2$). In addition, we performed a statistical significance test using the Wilcoxon signed-rank test to compare the performance of our framework with other comparative methods. The results, as presented in TABLE \ref{table7}, demonstrated that our proposed method achieved the highest MAPE and $R^2$ for all features. This outcome indicates the superior performance of our approach compared to the other methods under consideration. 
% Despite the fact that our proposed method is primarily focused on classification challenges, the experiment results show that the proposed method can be extended to different time series tasks.}

\subsubsection{Irregularly-Sampled Data (CharacterTrajectories)}
To show the effectiveness of our model for continuous models \cite{kidger2020neural, rubanova2019latent, de2019gru} dealing with irregularly-sampled time-series data, we conducted additional experiments using a CharacterTrajectories dataset from the UEA archive. To do this, we randomly dropped 30\%, 50\%, and 70\% of time-series data across channels by following \cite{kidger2020neural}. While training our proposed model with irregular settings, we set the model structure to be the same as with regular settings. However, it was difficult to use the time window due to random missing values, so we calculated the SPD matrix for each time point using feature maps obtained via 3 convolutional layers with a kernel size of 1 and batch normalization, LeakyReLU activation. 
% In general, the SPD matrix in multivariate time-series data indicates the correlation between channels over time. This makes it scalable representation learning with efficient computation cost. Unfortunately, in the case of irregularities, the singular matrix which does not satisfy the SPD matrix assumption is occurred due to missing values. Therefore, we considered the correlation between channels by calculating the feature map at each time point through the encoder.

We evaluated the mean and variance of accuracy through five repeats and reported in TABLE \ref{table6}. Our proposed method showed the same or higher performance at 30\% and 70\% drops compared with the baseline methods. Although it showed 0.5 lower performance in 50\% dropped than Neural CDE \cite{kidger2020neural}, it was still an overwhelming performance compared to other continuous models \cite{rubanova2019latent, de2019gru}. This indicates that our proposed model encouraged sufficient improvement even on irregularly-sampled data.

\subsubsection{Ablation Study}
To verify the effectiveness of the manifold ODE in our method, we conducted ablation studies by taking off the manifold ODEs. We conducted experiments on the SleepEDF-20 dataset, and the results are illustrated in TABLE \ref{table4}. Without manifold ODEs, ODE-RGRU (w/o ODEFunc.) showed a lower performance than ODE-RGRU. In particular, MF1 and sensitivity were the lowest among baseline models in both cases of EEG and EEG/EOG. This shows that time-series modeling has made a sufficient improvement in learning geometric characteristic representations through manifold ODE.

\begin{figure}
    \centering
    \includegraphics[width=1\linewidth]{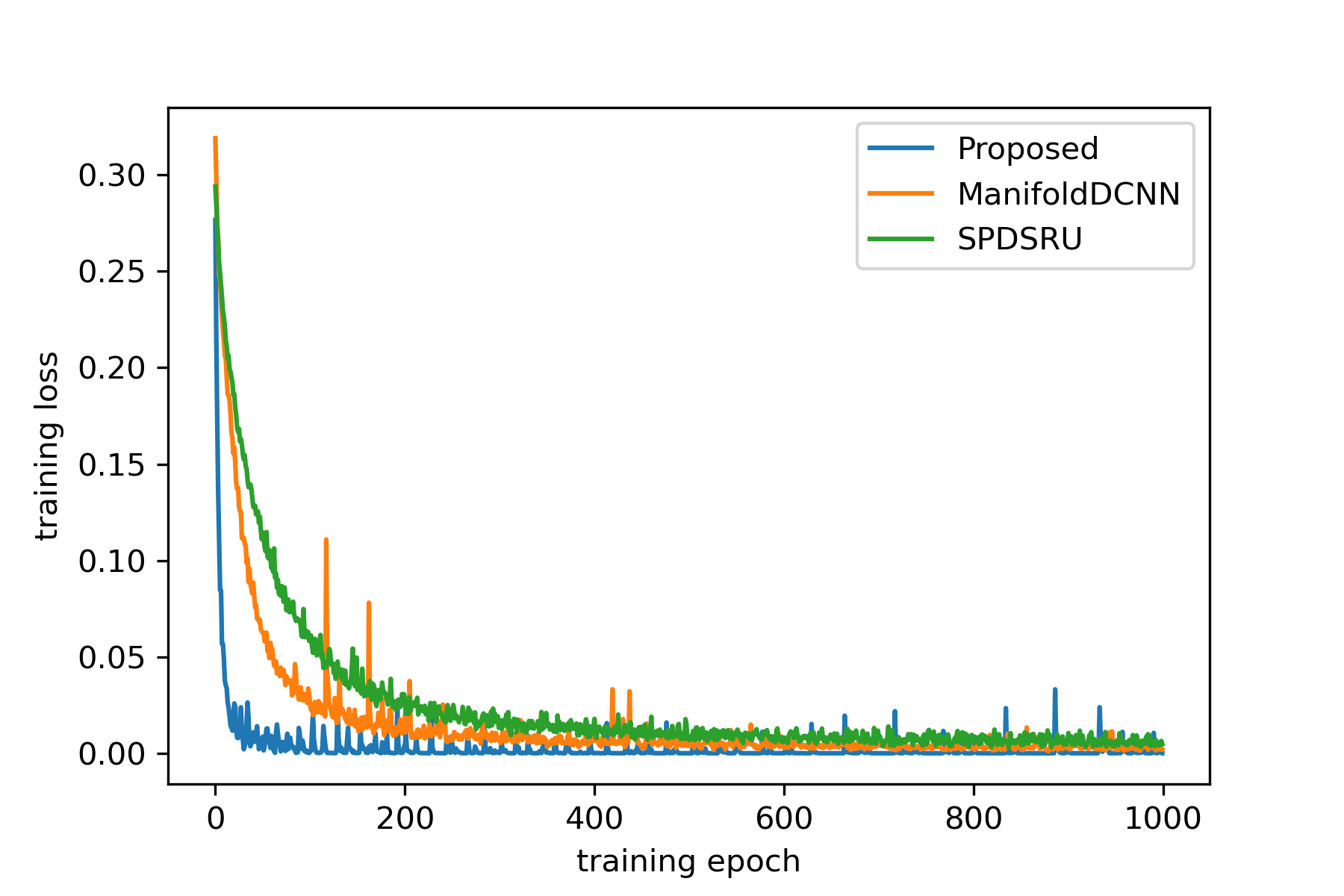}
    \caption{Comparison of the training loss curves of the manifold-based methods over the UCF11 dataset.}
    \label{fig3}
\end{figure}

\subsubsection{Training Speed}
Our method uses Cholesky decomposition to map the SPD manifolds into Cholesky space. The Cholesky decomposition greatly reduces the computational costs, allowing us to use the larger SPD matrix than the existing geometric deep-learning methods \cite{huang2017riemannian, brooks2019riemannian}. Therefore, by using the large SPD matrix, the proposed method could learn various features of the large dimensional input data. We compared the learning curves of three comparative manifold deep-learning methods in Fig. \ref{fig3}. Remarkably, our proposed method showed its convergence faster and more stable than the others.

\section{Conclusion}
\label{sec:conclusion}
Manifold learning with deep neural networks for SPD matrices is a challenge due to their hard constraints. In this paper, we proposed a novel deep efficient continuous manifold neural network, called ODE-RGRU, by systematically combining GRU and ODE in Cholesky space, and efficiently resolving difficulties related to computational costs in optimization. Specifically, we first reformulated the gating operations in GRU, thus making it work in the Riemannian manifold. Furthermore, to learn continuous geometric feature representations from sequential SPD-valued data, we additionally devised and applied a manifold ODE to the recurrent model.  
% Also, we developed a recurrent model in Cholesky space called RGRU by defining operations satisfying Cholesky properties.
% By applying manifold ODEs to the recurrent model, continuous geometric features can be extracted from sequential-valued data.
In our extensive experiments on various time-series tasks, we empirically demonstrated its effectiveness and suitability for time-series representation modeling, by achieving state-of-the-art performance for classification, forecasting, and imputation tasks.

\ifCLASSOPTIONcompsoc
  \section*{Acknowledgments}
\else
  \section*{Acknowledgment}
\fi
This work was supported by Institute of Information \& communications Technology Planning \& Evaluation (IITP) grant funded by the Korea government (MSIT) (No. 2022-0-00959; (Part 2) Few-Shot Learning of
Causal Inference in Vision and Language for Decision Making, No.
2019-0-00079; Artificial Intelligence Graduate School Program (Korea
University), and No. 2017-0-00451; Development of BCI based Brain and Cognitive Computing Technology for Recognizing User’s Intentions using Deep Learning).

\ifCLASSOPTIONcaptionsoff
  \newpage
\fi

\bibliographystyle{IEEEtran}
\bibliography{main}

\begin{IEEEbiography}[{\includegraphics[width=1in,height=1.25in,clip,keepaspectratio]{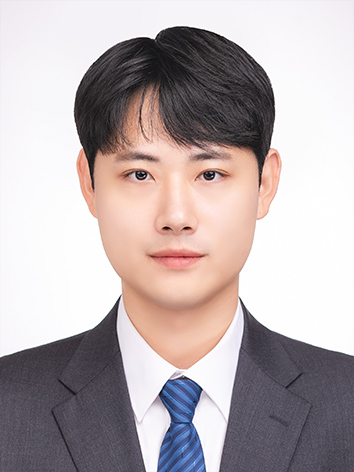}}]{Seungwoo Jeong}
	received the B.S. degree in Mathematics and Statistics from Hankuk University of Foreign Studies, Yongin, South Korea, in 2019. He is currently pursuing the Ph.D. degree with the Department of Artificial Intelligence, Korea University, Seoul, South Korea.
    
    His current research interests include time-series analysis, brain-computer interface, and machine/deep learning. 
\end{IEEEbiography}

\begin{IEEEbiography}[{\includegraphics[width=1in,height=1.25in, trim={30cm 0 5cm 0}, clip, keepaspectratio]{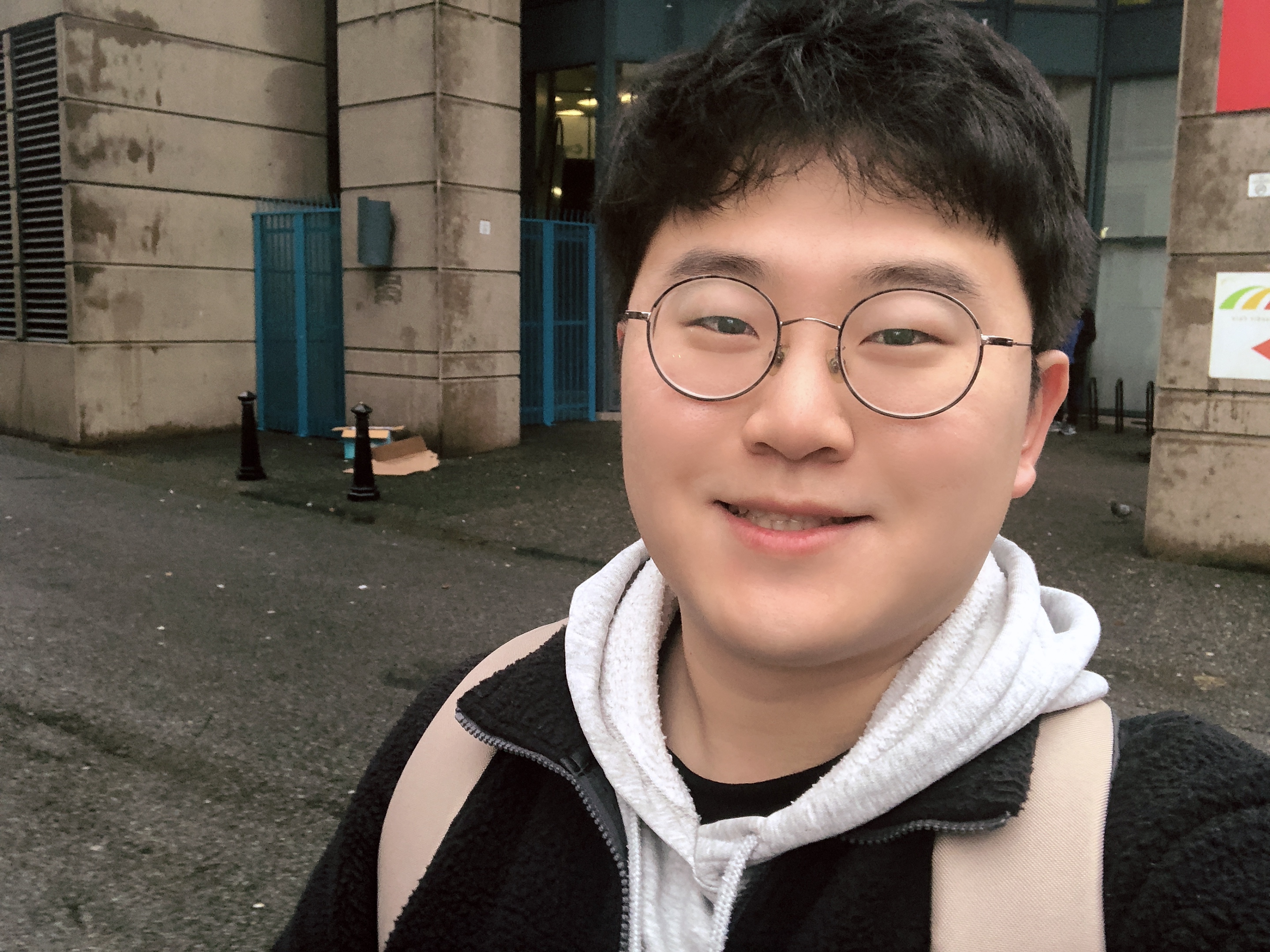}}]{Wonjun Ko}
    received the Ph.D. degree in brain and cognitive engineering from Korea University, Seoul, Republic of Korea, in 2022. He was a Data Scientist with SK Hynix Inc.,  Incheon, Republic of Korea, from 2022 to 2023.

    He is currently an Assistant Professor at the School of AI Convergence, Sungshin Women's University, Seoul, Republic of Korea. His current research interests include machine learning, representation learning, and data mining.
\end{IEEEbiography}

\begin{IEEEbiography}[{\includegraphics[width=1in,height=1.25in,clip,keepaspectratio]{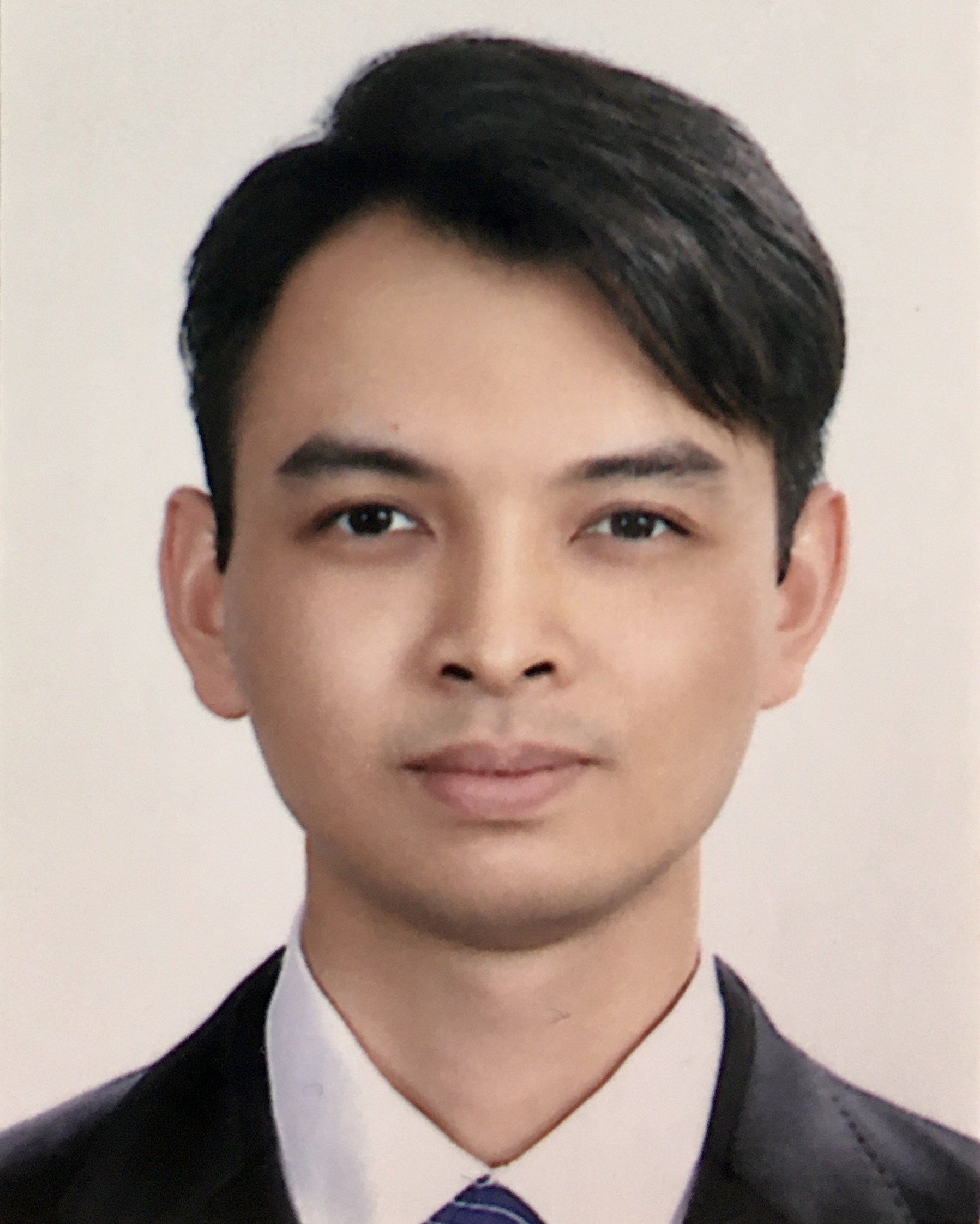}}]{Ahmad Wisnu Mulyadi} 
	received the bachelor’s degree in computer science education from the Indonesia University of Education, Bandung, Indonesia,  in 2010.  He is currently pursuing the Ph.D. degree with the Department of Brain and Cognitive Engineering, Korea University, Seoul, South Korea.

His current research interests include machine/deep learning in healthcare, biomedical image analysis, and graph representation learning.
\end{IEEEbiography}

\begin{IEEEbiography}[{\includegraphics[width=1in,height=1.25in,clip,keepaspectratio]{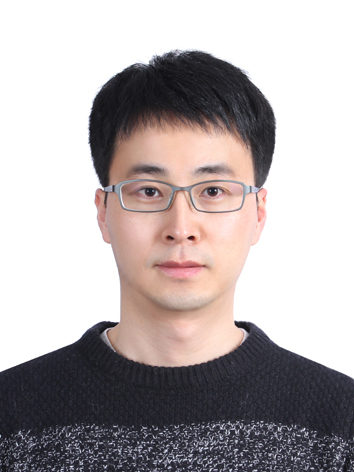}}]{Heung-Il Suk}
     is currently a Professor at the Department of Artificial Intelligence and an Adjunct Professor at the Department of Brain and Cognitive Engineering at Korea University. He was a Visiting Professor at the Department of Radiology at Duke University between 2022 and 2023.

     He was awarded a Kakao Faculty Fellowship from Kakao and a Young Researcher Award from the Korean Society for Human Brain Mapping (KHBM) in 2018 and 2019, respectively. His research interests include causal machine/deep learning, explainable AI, biomedical data analysis, and brain-computer interface.

     Dr. Suk serves as an Editorial Board Member for Clinical and Molecular Hepatology (Artificial Intelligence Sector), Electronics, Frontiers in Neuroscience, Frontiers in Radiology (Artificial Intelligence in Radiology), International Journal of Imaging Systems and Technology (IJIST), and a Program Committee or a Reviewer for NeurIPS, ICML, ICLR, AAAI, IJCAI, CVPR, MICCAI, AISTATS, \etc.

\end{IEEEbiography}

\end{document}